\documentclass[10pt,twocolumn,letterpaper]{article}

\usepackage[pagenumbers]{cvpr} % To force page numbers, e.g. for an arXiv version

\usepackage[utf8]{inputenc} % allow utf-8 input
\usepackage[T1]{fontenc}    % use 8-bit T1 fonts

\usepackage{url}            % simple URL typesetting
\usepackage{booktabs}       % professional-quality tables
\usepackage{amsfonts}       % blackboard math symbols
\usepackage{nicefrac}       % compact symbols for 1/2, etc.
\usepackage{microtype}      % microtypography
\usepackage{xcolor}         % colors

\definecolor{cvprblue}{rgb}{0.21,0.49,0.74}
\usepackage[pagebackref,breaklinks,colorlinks,allcolors=cvprblue]{hyperref}
\usepackage{tablefootnote}

\usepackage{multirow}
\usepackage{xspace}
\usepackage{amsmath}
\usepackage[table]{xcolor}
\usepackage{subcaption}
\usepackage{enumitem}
\usepackage{algorithm}
\usepackage{wrapfig}
\usepackage{minitoc}

\usepackage[toc,page]{appendix}
\usepackage{fancyhdr}
\usepackage[titles]{tocloft}
\usepackage{titletoc}
\usepackage{etoolbox}

\usepackage[noend]{algcompatible}
\newcommand{\methodname}{\textsc{ICONS}\xspace}
\newcommand{\mini}{\textsc{LLaVA-ICONS-133K}\xspace}
\newcommand{\full}{\textsc{LLaVA-665K}\xspace}
\newcommand{\cambrian}{\textsc{Cambrian-7M}\xspace}
\newcommand{\cambrianmini}{\textsc{Cambrian-ICONS-1.4M}\xspace}
\newcommand{\visionflan}{\textsc{Vision-Flan-186K}\xspace}
\newcommand{\visionflanmini}{\textsc{Vision-Flan-ICONS-37K}\xspace}

\DeclareMathOperator*{\argmax}{arg\,max}
\newcommand{\PAR}[1]{\vskip4pt \noindent {\bf #1~}}

\newcommand{\relp}{Rel.\xspace}

\usepackage[framemethod=TikZ]{mdframed}
\usepackage{tcolorbox}
\definecolor{chart}{HTML}{1f77b4}

\newtcolorbox{example}[1][]{
  colback=chart!5!white,
  colframe=chart,
  floatplacement=floating,
  title=\centering \textsf{#1}
}

\usepackage[framemethod=TikZ]{mdframed}
\usepackage{tcolorbox}
\definecolor{chart}{HTML}{1f77b4}
\mdfsetup{% 
    middlelinecolor =   none,
    middlelinewidth =   1pt,
    backgroundcolor =   blue!5,
    roundcorner     =   5pt,
}

\definecolor{coco1}{HTML}{D9E4EC}
\definecolor{coco2}{HTML}{B7CFDC}
\definecolor{coco3}{HTML}{6AABD2}
\definecolor{coco4}{HTML}{385E72}
\hypersetup{
    colorlinks=true,
    linkcolor=coco3,
    filecolor=coco4,      
    urlcolor=coco3,
    citecolor=coco3,
}
\usepackage{amsmath,amsfonts,bm,bbm}

\newcommand{\RETURN}{\STATE \textbf{return} }

\definecolor{Gray}{gray}{0.9}

\def\vtheta{{\bm{\theta}}}

\def\vx{{\bm{x}}}
\def\vy{{\bm{y}}}
\def\vz{{\bm{z}}}
\def\vI{{\bm{I}}}

\DeclareMathAlphabet{\mathsfit}{\encodingdefault}{\sfdefault}{m}{sl}
\SetMathAlphabet{\mathsfit}{bold}{\encodingdefault}{\sfdefault}{bx}{n}

\def\paperID{} % *** Enter the Paper ID here
\def\confName{CVPR}
\def\confYear{2026}

\title{ICONS: Influence Consensus for Vision-Language Data Selection}

\author{
    Xindi Wu$^{1}$
    \qquad 
    Mengzhou Xia$^{1}$
    \qquad 
    Rulin Shao$^{2}$ 
    \qquad 
    Zhiwei Deng$^{3}$ \\
    \qquad 
    Pang Wei Koh$^{2,4}$ 
    \qquad
    Olga Russakovsky$^{1}$
    \\
    \small
    $^{1}$Princeton University
    \quad 
    $^{2}$University of Washington 
    \quad 
    $^{3}$Google DeepMind
    \quad 
    $^{4}$Allen Institute for AI
\\
\small{\texttt{\url{https://princetonvisualai.github.io/icons/}}}
}

\begin{document}

\maketitle

\begin{abstract}
Training vision-language models via instruction tuning relies on large data mixtures spanning diverse tasks and domains, yet these mixtures frequently include redundant information that increases computational costs without proportional gains. 
Existing methods typically rely on task-agnostic heuristics to estimate data importance, limiting their effectiveness across tasks.
We introduce \methodname, a gradient-based \underline{I}nfluence \underline{CON}sensus approach for vision-language data \underline{S}election. 
Our method leverages first-order training dynamics to estimate each example's influence on validation performance, then aggregates these estimates across tasks via majority voting. This cross-task consensus identifies consistently valuable data points while mitigating score calibration and outlier sensitivity, enabling robust and scalable data selection for diverse multitask mixtures.
Models trained on our selected 20\% data subset from \full (respectively: from \cambrian, from \visionflan) retain 98.6\% (respectively: 98.8\%, 99.8\%) of full-dataset performance. 
We demonstrate that our selected data generalizes to unseen tasks and model architectures, and release three compact subsets \mini, \cambrianmini, and \visionflanmini for efficient vision-language model development.
\end{abstract}

\section{Introduction}
\label{sec:intro}
% %\vspace{-5pt} 

\begin{figure}[t]
    \centering
    \includegraphics[width=\linewidth]{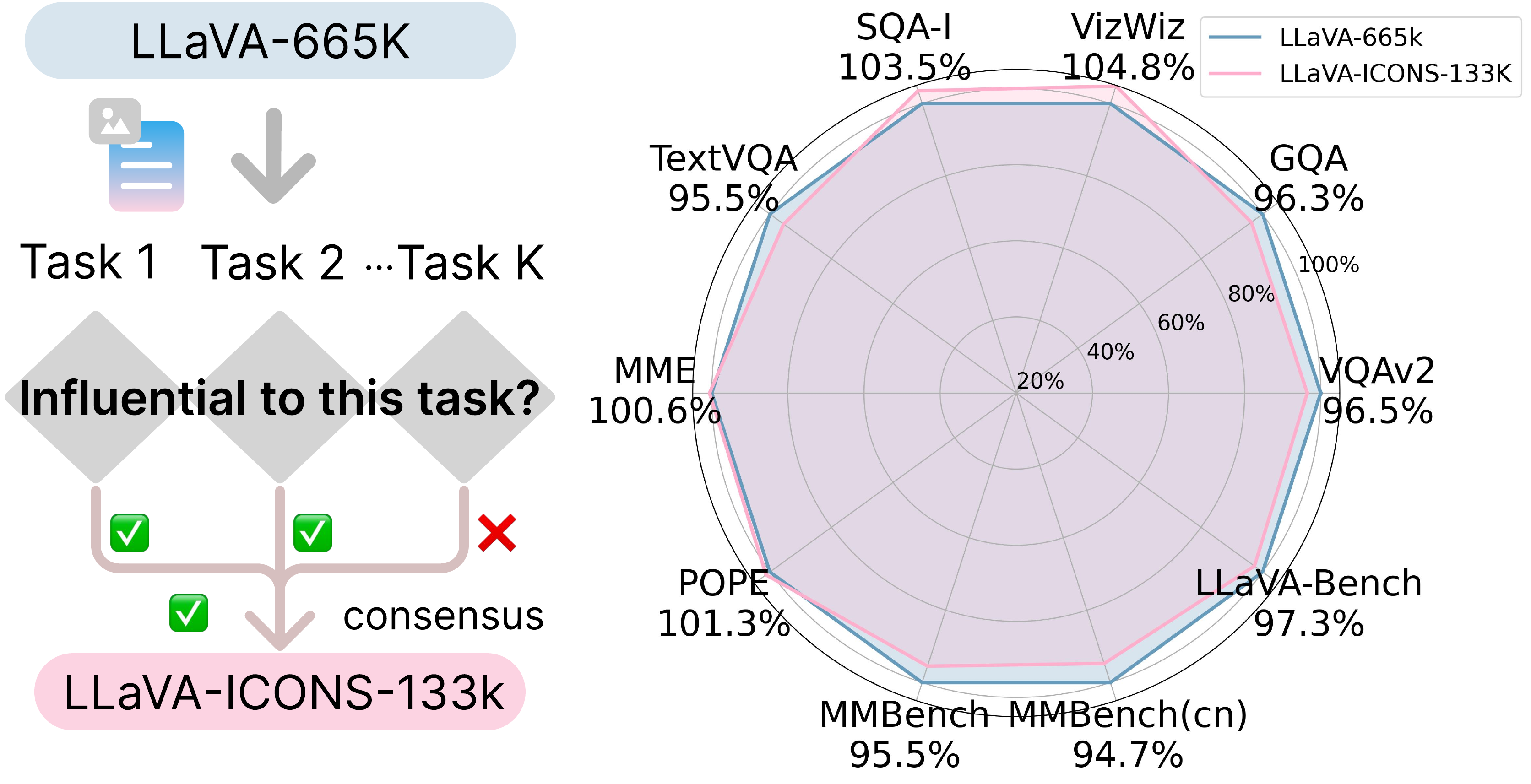}
    \caption{\textbf{Influence consensus for vision-language data selection}. 
    (\textit{Left}) Given a large-scale visual instruction tuning dataset (\full), 
    our method uses majority voting across task-specific influence scores to identify training samples that are consistently influential across multiple tasks, forming a compact 20\% subset (\mini) with data points achieving influence consensus.
    (\textit{Right}) The radar plot compares performance between \full and our selected subset, showing the selected subset achieves comparable results to the full dataset.}
    \label{fig:teaser}
\end{figure}

Visual instruction tuning (VIT) is a crucial step in training vision-language models (VLMs)~\cite{liu2024improved, llava}, enabling them to follow language instructions grounded in visual content. Recent approaches rely on large-scale instruction following datasets such as \full~\cite{liu2024improved} and \cambrian~\cite{tong2024cambrian}. While effective, these datasets introduce significant barriers to model training and deployment: prolonged training times~\cite{brown2020language, kaplan2020scaling}, high storage demands~\cite{rajbhandari2020zero, gadre2024datacomp}, and substantial compute costs~\cite{maini2024rephrasing, wu2022sustainable}. Moreover, not all training samples contribute equally to diverse downstream tasks and naively scaling up diverse data mixtures can introduce redundancy and inefficiency. This raises a fundamental question:

\begin{mdframed}[leftmargin=0pt, rightmargin=0pt, innerleftmargin=10pt, innerrightmargin=10pt, skipbelow=0pt]
\begin{center}
\emph{Can we identify a compact, multitask-effective subset of training data that preserves model capabilities across tasks while enabling faster experimentation?}
\end{center}
\end{mdframed}

Prior work has explored various data selection strategies, including gradient-based approaches~\cite{xia2024less, deng2024influential}, influence functions~\cite{yang2022dataset, koh2017understanding}, and diversity-based sampling~\cite{yu2024diversify, bukharin2023data}. However, many of these methods either optimize for single tasks in isolation or maximize source diversity without aligning to downstream needs. In the context of VIT data that supports diverse vision-language tasks, this is particularly limiting: optimizing for one task may hurt generalization, and task-agnostic diversity may dilute impact. Rather than selecting data based on per-task influence, we aim to identify samples that are broadly useful, with training examples that consistently contribute across tasks. To do this, we aggregate gradient-based influence scores using a simple yet effective majority voting approach to reach influence consensus.

We introduce \methodname (\underline{I}nfluence \underline{CON}sensus vision-language data \underline{S}election), a method that builds upon LESS~\cite{xia2024less}, a gradient-based selection approach. 
Given access to validation data for each target task, our method: (1) computes first-order gradient influence scores to measure how each training sample impacts task-specific performance, and (2) reaches influence consensus via majority voting to identify training samples that show consistent positive impact across multiple tasks. This consensus-based mechanism identifies universally valuable training examples: while some samples might be highly influential for individual tasks, we prioritize those that demonstrate broad utility across the task spectrum. While the computational cost of influence estimation is high, this front-loaded, one-time investment yields a standardized, compact dataset that can significantly accelerate development of multimodal models, and enables reusable gradient datastores that amortize costs across iterations and deliver long-term savings. 

Using \methodname, we create \mini, \cambrianmini, and \visionflanmini, automatically selected 20\% subsets of \full~\cite{liu2024improved}, \cambrian~\cite{tong2024cambrian}, and \visionflan~\cite{visionflan} datasets, respectively. These compact datasets maintain 98.6\%, 98.8\%, and 99.8\% of their original performance across multiple vision-language tasks, providing significant improvements over randomly selecting same-sized subsets (95.8\%, 95.4\%, and 91.6\%) and eliminating approximately two-thirds of the performance drop from shrinking the training data. Moreover, our \methodname outperforms all baselines across different selection ratios, and remarkably achieves above-full-dataset performance, surpassing the original datasets at a 60\% selection ratio for \full. Importantly, the selected subsets show strong transferability, e.g., \mini maintains 95.5-113.9\% relative performance across unseen tasks, suggesting that \methodname identifies fundamentally valuable training data. We summarize our key contributions: 
\begin{enumerate}
\item We propose \methodname, a simple yet effective method for multitask vision-language data selection that identifies broadly valuable training examples via majority voting over task-specific gradient influence scores.
\item Our consensus-based selection outperforms all baselines (\S\ref{subsec:baseline}) and we ablate influence aggregation strategies and show the advantage of voting-based consensus (\S\ref{subsec:ablation}). \methodname exceeds 102\% of full-dataset performance at a 60\% selection ratio on \full (\S\ref{subsec:ratio}).
\item We release subsets \mini, \cambrianmini and \visionflanmini for resource-efficient development, demonstrating competitive performance including generalization to unseen tasks (\S\ref{subsec:transferability}) despite a 5x reduction in training data. 
\end{enumerate}

\section{Related work}
\label{sec:related_work}
%\subsection{Visual instruction tuning}
\paragraph{Visual instruction tuning.} Vision-language models (VLMs), e.g., Flamingo~\cite{flamingo}, LLaVA~\cite{llava}, BLIP2~\cite{blip-2}, and Cambrian~\citep{tong2024cambrian}, enable understanding and reasoning across visual and textual modalities for various multimodal tasks.
A key component in advancing VLMs is visual instruction tuning (VIT)~\citep{llava}, a training process that enables these models to interpret and follow instructions within a vision-language context, transforming them into versatile multimodal assistants. 
This tuning process not only improves the models' instruction-following abilities but also aligns their outputs more closely with user expectations, thus enhancing their utility in practical applications~\citep{llava}. 
To support this paradigm, recent VIT datasets have scaled to contain hundreds of thousands to millions of instruction-response pairs~\cite{liu2024improved,tong2024cambrian}.

%\subsection{Data selection}
\paragraph{Data selection.} Data selection methods~\cite{survey} can be categorized based on the types of information they utilize for selection. Representation-based approaches~\citep{semdedup,coincide} leverage embeddings to capture data representations.
Loss trajectory-based methods~\citep{mindermann2022prioritized} prioritize data points that contribute most significantly to reducing generalization error over training. 
Gradient-based techniques~\citep{el2n,xia2024less,deng2024influential} select data based on gradient information. 
Recent work has explored various approaches to select optimal visual instruction tuning datasets. Concurrent work TIVE~\citep{liu2024less} employs gradient-based selection to identify representative instances. TIVE assumes that the number of specialist data should be proportional to task difficulty and thus samples specialist data based on an estimation of task difficulty. Our method does not rely on this assumption -- we directly select samples that benefit the greatest number of tasks. 
COINCIDE~\citep{coincide} clusters data based on representations associated with concept-skill compositions. 
Our work follows targeted instruction tuning selection approach LESS~\cite{xia2024less} to utilize gradient information to calculate the specialist influence (i.e., the influence on a specific task) and extends it to general scenarios by aggregating information across tasks and selecting data influential for multiple downstream tasks via majority voting.

\section{\methodname: \underline{I}nfluence \underline{CON}sensus for vision-language data \underline{S}election}
\label{sec:method}

\begin{figure*}[t!]
 % %\vspace{-10mm}
  \centering
  \includegraphics[width=\linewidth]{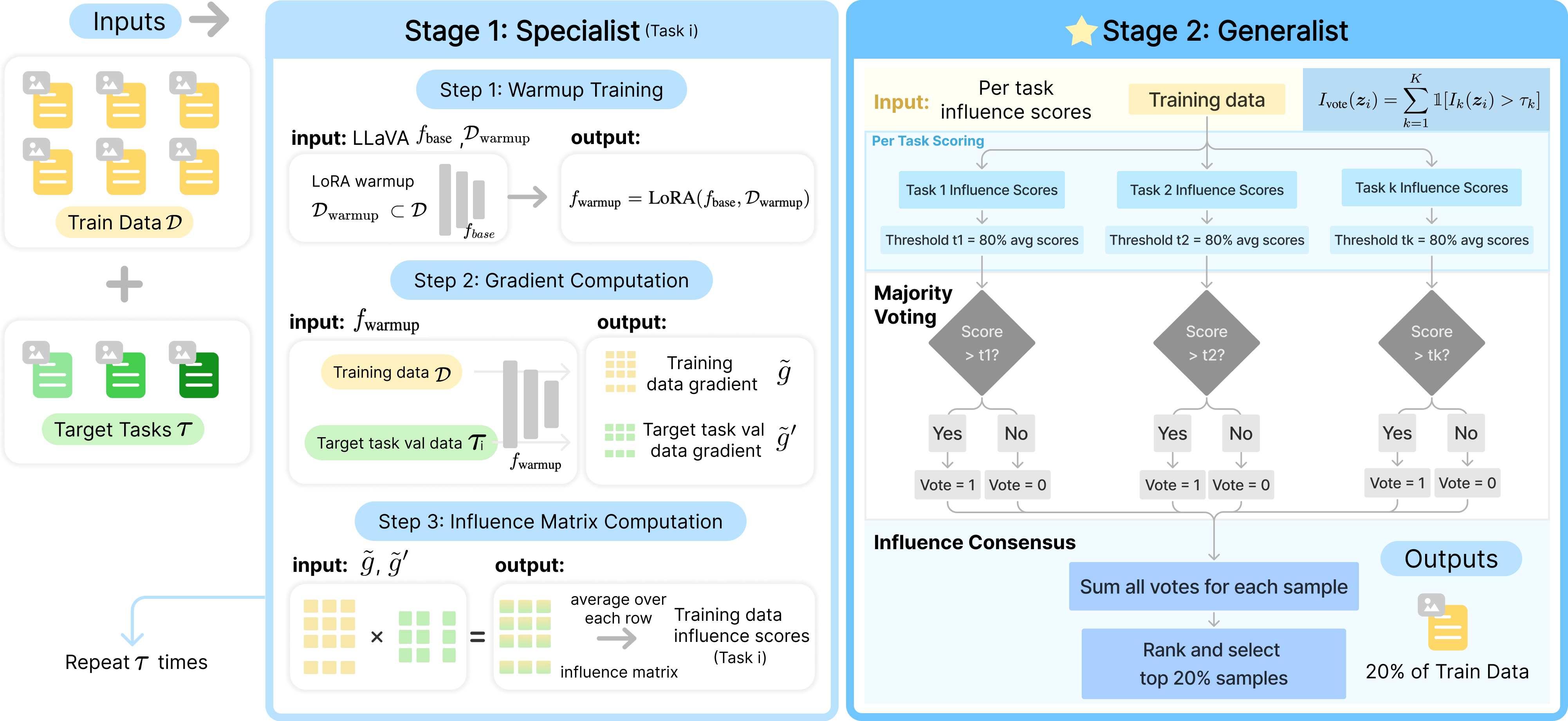}
  %\vspace{-10pt}
   \caption{\textbf{\methodname.} 
   The Specialist stage (\textit{left}) processes each task through: (1) warmup training on a small subset, (2) gradient computation for training and validation data, and (3) influence matrix computation for per-task scores. 
   The Generalist stage (\textit{right}) performs \textbf{Influence Consensus} to aggregate information across tasks. For each task, samples scoring above a percentile threshold (e.g., $80^{\text{th}}$ for 20\% selection) receive a vote. Votes are summed across tasks, and samples with the highest vote counts are selected, creating a compact yet effective training dataset that performs well across all tasks. 
   } 
   \vspace{-10pt}
  \label{fig:pipeline}
\end{figure*}

We propose a consensus-driven, gradient-based data selection framework (Fig.~\ref{fig:pipeline}) for visual instruction tuning datasets. We formalize the problem setup in \S\ref{subsec:problem} and establish gradient-based influence estimation preliminaries in \S\ref{subsec:preliminary}. Our two-stage data selection framework consists of: the \textit{specialist} stage (\S\ref{subsec:specialist}), which computes task-specific influence scores, and the \textit{generalist} stage (\S\ref{subsec:generalist}), which builds cross-task consensus through voting-based aggregation.

\subsection{Problem formulation}
\label{subsec:problem}
Given a large-scale visual instruction tuning dataset $\mathcal{D} = \{(\vI_i, \vx_i, \vy_i)\}_{i=1}^N$ containing $N$ samples, where each data point $\vz_i = (\vI_i, \vx_i, \vy_i)$ includes an image $\vI_i$, natural language instruction $\vx_i$, and corresponding target response $\vy_i$\footnote{This framework supports multi-turn conversational data, yet we formalize the setup for single-turn instruction-tuning for clarity and simplicity.}, 
and given access to validation data $\mathcal{V}_k$ for each downstream task $T_k \in \mathcal{T} = \{T_1, ..., T_K\}$,
our goal is to select a compact subset $\mathcal{S} \subset \mathcal{D}$ of size $M \ll N$ that maximizes model performance across multiple downstream tasks. 

Concretely, we let $f_{\mathcal{S}}$ and $f_{\mathcal{D}}$ denote models trained on subset $\mathcal{S}$ and full dataset $\mathcal{D}$, respectively, and $\text{Score}(f, T_k)$ be the task-specific evaluation score achieved by model $f$ on task $T_k$. $\text{Rel}(f_{\mathcal{S}}, T_k) = \frac{\text{Score}(f_{\mathcal{S}}, T_k)}{\text{Score}(f_{\mathcal{D}}, T_k)}$ quantifies the subset-trained model's performance relative to that of the model trained on the entire dataset, with values close to 1 indicating that the subset maintains the performance of full training~\cite{coincide}. Our goal is then to identify the subset:
%\vspace{-7pt}
\begin{equation}
\mathcal{S}^* = \argmax_{\mathcal{S} \subset \mathcal{D}, |\mathcal{S}| = M} \sum_{k=1}^K 
\text{Rel}(f_{\mathcal{S}}, T_k).
\label{eq:rel}
\end{equation}
% %\vspace{-5pt}

We define the average relative performance across all tasks as \textbf{Rel.} = $\sum_{k=1}^K \text{Rel}(f_{\mathcal{S}}, T_k) / K$, and our objective is to select a subset where Rel. $\approx 1$, i.e., the model trained on the subset achieves comparable performance to the one trained with full dataset.

\subsection{Preliminaries}
\label{subsec:preliminary}
Building on the problem formulation in \S\ref{subsec:problem}, we formalize how to estimate the influence of training samples on downstream task performance. 
Since our goal is to maximize $\text{Rel}(f_{\mathcal{S}}, T_k)$ across tasks as defined in Eqn.~\ref{eq:rel}, we need an efficient way to estimate how each training sample contributes to the $\text{Score}(f_{\mathcal{S}}, T_k)$ term in the numerator.
Denote a training data point as $\vz$ and a validation data point as $\vz'$ from validation set $\mathcal{V}_k$ for task $T_k$. Following~\cite{pruthi2020estimating, xia2024less}, we estimate how $\vz \in \mathcal{D}$ affects validation loss by measuring its gradient alignment with reducing validation loss on $\mathcal{V}_k$, which directly impacts task-specific evaluation.
When training with SGD and batch size 1, using data point $\vz$ at timestep $t$ leads to a model update $\vtheta_{t+1} = \vtheta_t - \eta_t \nabla\ell(\vz;\vtheta_t)$, where $\eta_t$ is the learning rate. 
To reduce the computational cost, we use the first-order Taylor expansion to estimate the loss on a given validation data point $\vz'$ at time step $t+1$ as:
\begin{equation*}
	\ell(\vz';\vtheta_{t+1}) \approx \ell(\vz';\vtheta_t) + \langle\nabla\ell(\vz';\vtheta_t), \vtheta_{t+1}-\vtheta_t\rangle.
\end{equation*}
Training sample $\vz$'s influence on validation sample $\vz'$ is:
\begin{align*}
    \mathcal{I}_t(\vz \rightarrow \vz') &= \ell(\vz';\vtheta_{t+1}) - \ell(\vz';\vtheta_t) \\
    &\approx -\eta_t\langle\nabla\ell(\vz';\vtheta_t), \nabla\ell(\vz;\vtheta_t)\rangle,
\end{align*}
which we refer to as an influence score.
The gradient-based selection approach selects training samples $\{\vz\}$ that maximize the gradient inner product $\langle\nabla\ell(\vz';\vtheta_t), \nabla\ell(\vz;\vtheta_t)\rangle$\footnote{In practice, we use cosine similarity instead of direct inner products to avoid biasing selection toward shorter sequences, since gradient norms tend to be inversely correlated with sequence length as noted in~\cite{xia2024less}.} through a greedy, first-order approximation, which leads to larger reductions in validation loss for point $\vz'$.
While it omits second-order terms compared to influence functions~\cite{koh2017understanding}, it provides an efficient approximation to rank the impact of training samples~\cite{xia2024less}.

\subsection{Specialist: individual task influence ranking }
\label{subsec:specialist}

To rank the influence of training data for each target task, we compute the influence score of each training data point on a validation set that represents the target task distribution. Following LESS~\cite{xia2024less}, the process involves three steps: (1) training the model on 5\% randomly selected data as a lightweight warm-up to initialize visual instruction-following capabilities, (2) computing gradients for training and validation data and compressing the gradients via random projection, and (3) computing the influence score to quantify the impact of each training data on validation set.

\PAR{Step 1: Warm-up Training.}
Following LESS~\cite{xia2024less}, we apply LoRA~\cite{hu2021lora} on a small random subset $\mathcal{D}_\mathrm{warmup} \subset \mathcal{D}$ (5\%) to obtain $f_{\text{warmup}} = \text{LoRA}(f_{\text{base}}, \mathcal{D}_{\text{warmup}})$, which develops basic visual instruction-following capabilities.

\PAR{Step 2: Gradient computation.}
For each training data $\vz_i \in \mathcal{D}$ and validation data $\vz'_j \in \mathcal{D}_{\text{val}}^k$ from $\mathcal{T}_k$, we compute their gradients with respect to $f_{\text{warmup}}$ parameters $\theta_{w}$:
\begin{equation*}
    g_i = \nabla_{\theta_w} \mathcal{L}(f_{\text{warmup}}(\vz_i), \vy_i), \quad
    g'_j = \nabla_{\theta_w} \mathcal{L}(f_{\text{warmup}}(\vz'_j), \vy'_j)
\end{equation*}
where $y_i$ and $y'_j$ are the targets for $z_i$ and $z'_j$, respectively.
In order to reduce computational and storage overhead, we apply random projection to the gradient feature: $\tilde{g}_i = R g_i$ and $\tilde{g}'_j = R g'_j$, where $R \in \mathbb{R}^{d' \times d}$ is a random projection matrix with $d' \ll d$ that preserves inner products with high probability \citep{johnson1984extensions}. We further normalize the projected gradients, 
$\tilde{g}_i = \frac{\tilde{g}_i}{\|\tilde{g}_i\|_2}, 
\tilde{g}'_j = \frac{\tilde{g}'_j}{\|\tilde{g}'_j\|_2}$
to prevent bias from sequence length differences~\cite{xia2024less}.

\PAR{Step 3: Influence matrix computation.}
We compute the influence matrix $I \in \mathbb{R}^{|\mathcal{D}| \times |\mathcal{D}_{\text{val}}^k|}$ where each entry $I_{ij} = \langle \tilde{g}_i, \tilde{g}'_j \rangle$ is the influence of the training data $\vz_i$ on the validation data $\vz'_j$, 
following the setting in LESS~\cite{xia2024less}, we take the average influence of training data $\vz_i$ across all validation examples on the target task $k$ to obtain a representative task-level influence score: 
% %\vspace{-5pt}
\begin{equation}
    \bar{I}_{k}(\vz_i) = \frac{1}{|\mathcal{D}_{\text{val}}^k|} \sum_{j=1}^{|\mathcal{D}_{\text{val}}^k|} I_{ij}.
\label{eqn:influence_matrix}
\end{equation}
This influence estimation process provides a task-specific ranking for the training set $\mathcal{D}$ with respect to task $\mathcal{T}_k$, where a higher influence score $\bar{I}_i$ suggests a higher influence on $\mathcal{T}_k$.
We can select a small training subset $\mathcal{S}_k$ for a given task $k$ by selecting the training examples $\vz_i$ with the highest-influence values $\bar{I}_{k}(\vz_i)$. This simple greedy approach has been shown by LESS to be successful, and thus we use it as our task-specific (``specialist'') baseline. However, recall that our goal is to select a single compact subset that maximizes the performance across \emph{all} tasks. We address this disconnection between task-specific rankings and overall optimization by proposing a voting-based generalist approach to identify the most broadly impactful training data.

\subsection{Generalist: cross-task influence consensus} 
\label{subsec:generalist}

Our goal is to identify a training set subset $\mathcal{S} \subset \mathcal{D}$ of size $M \ll N$ such that its performance across all tasks remains high as defined by Eqn.~\ref{eq:rel}. There are multiple ways to tackle this, depending on the assumptions one makes about the task-specific influence scores $\bar{I}_{k}(\vz_i)$. The simplest approach is to merge together all the different tasks' validation sets $\mathcal{D}_{\text{val}}^k$ (normalizing for their different sizes) and compute the total influence score for a training example $\vz_i$ as: 
% %\vspace{-7pt}
\begin{equation}
    I_{\text{Merge}}(\vz_i) = \sum_{k=1}^{K}\bar{I}_{k}(\vz_i).
\label{eqn:merge}
\end{equation}
LESS~\cite{xia2024less} suggests a similar approach:
\begin{equation}
    I_{\text{Max}}(\vz_i) = \max_{k=1,\dots K}\bar{I}_{k}(\vz_i),
\label{eqn:max}
\end{equation}
i.e., the influence of the data is measured as its \emph{highest} influence on any tasks. The set of $M$ training examples with the highest aggregated influence scores would be selected for inclusion in the training set $\mathcal{S}_\text{merge}$ (correspondingly, $\mathcal{S}_\text{max}$). Both approaches, however, require that the influence scores $\bar{I}_{k}$ be well-calibrated across the different tasks; as we show in \S\ref{subsec:ablation} this may not necessarily be the case.

An alternative approach which does not require directly comparing influence scores $\bar{I}_{k}$ across tasks $k$ is to leverage the relative rank of the training examples within each task. Concretely, we compute $\text{rank}_k(\vz_i)$ for each example $\vz_i$ relative to other examples for task $k$ according to their influence scores (higher influence scores correspond to lower rank). We can have a couple of options. First, we can select the training subset either using the Round Robin (RR) approach~\cite{ivison2025large} where we iterate over tasks and select the lowest-rank example which has not yet been selected to add to our set
$\mathcal{S}_\text{RR}$. Alternatively, we can select the training subset $\mathcal{S}_\text{MinRank}$ such that all the examples within it have a low rank for some task $k$. Mathematically, albeit somewhat confusingly, this corresponds to:
\begin{equation}
   \mathcal{S}_\text{MinRank} = \argmax_{\mathcal{S} \subset \mathcal{D}, |\mathcal{S}| = M} \min_{\substack{\text{task } k\\ \text{example }\vz_i \notin S}} \text{rank}_k(\vz_i ),
\label{eqn:minrank}
\end{equation}
i.e., all examples that are \emph{not} included in $\mathcal{S}$ would have high relative ranks $\text{rank}_k(\vz_i )$ for all tasks $k$.
However, this approach ignores the potential interplay between tasks. 
Since Eqn.~\ref{eq:rel} maximizes the \emph{sum} of relative performance across tasks $k$, if a training example is beneficial for multiple tasks, we may want to include it even at the expense of lower-ranked examples for individual tasks.
Thus, we introduce a consensus-based voting strategy that identifies training samples consistently showing high influence across tasks. 
For each task $k$, we set a threshold $\tau_k$ at the $((1-p) \times 100)$-th percentile of the influence score distribution, where $p$ represents the selection ratio ($p=0.2$ in our main experiments). 
A sample $\vz_i$ is considered important for task $k$ if $\bar{I}_k(\vz_i) \geq \tau_k$. We define the voting-based influence score as:
\vspace{-5pt}
\begin{equation}
I_{\text{vote}}(\vz_i) = \sum_{k=1}^{K} \mathbbm{1}[\bar{I}_k(\vz_i) \geq \tau_k],
\label{eq:vote}
\end{equation}
where $\mathbbm{1}[\cdot]$ is the indicator function. The final score ranges from 0 (no importance for any task) to $K$ (consistent importance for all tasks). Equivalently, we can define the threshold-based specialist sets as:
$\mathcal{S}_k = \{\vz_i \in \mathcal{D} : \bar{I}_k(\vz_i) \geq \tau_k\}$.
We then select the combined training set as:
\vspace{-5pt}
\begin{align}
\mathcal{S}_{\text{ICONS}} &= \argmax_{\mathcal{S} \subset \mathcal{D}, |\mathcal{S}| = M} \sum_{k=1}^K \sum_{\vz_i\in S}  \mathbbm{1}[\vz_i \in S_k] \\
&= \argmax_{\mathcal{S} \subset \mathcal{D}, |\mathcal{S}| = M} \sum_{\vz_i\in S} I_{\text{vote}}(\vz_i)
\end{align}
This simple approach offers a key advantage: it does not rely on calibration of influence scores across tasks, and does not make any a-priori assumptions about the relationship between tasks (e.g., that every task needs to have its highest-scoring examples included in the combined training set). 
Our generalist stage converts each task’s ranked list into a binary vote (``above threshold'' or not) and then combines these votes across tasks, eliminating the need for task-specific normalization. As a result, the selection remains insensitive to scale differences while still capturing relative importance within each task.
Meanwhile, a training sample is selected only when several tasks independently rank it as influential, preventing over-representation of single-task outliers and ensuring the cross-task utility.

\section{Experiments}
\label{sec:experiments}

We first discuss our experiment setup (\S\ref{subsec:exp_testbed}), then compare \methodname with state-of-the-art methods (\S\ref{subsec:baseline}) and analyze different selection strategies (\S\ref{subsec:ablation}). We evaluate transferability to unseen tasks (\S\ref{subsec:transferability}), performance across selection ratios (\S\ref{subsec:ratio}), vote distributions (\S\ref{app:vote_evolution}), and characteristics of low-vote samples (\S\ref{subsec:zero}).

\subsection{Evaluation test-bed}
\label{subsec:exp_testbed}
\PAR{Datasets \& model.}
We apply \methodname on major VIT training datasets: \full~\cite{liu2024improved}, \cambrian~\cite{tong2024cambrian} and \visionflan~\cite{visionflan}. The majority of our analysis and ablation experiments are conducted on \full.
For our experiments, we use the LLaVA-v1.5 model~\citep{liu2024improved} checkpoint after Stage 1 (pre-training for feature alignment) as defined in the original LLaVA training pipeline, with a default size of 7B parameters and \full unless otherwise specified.
This checkpoint\footnote{
\href{https://huggingface.co/liuhaotian/llava-v1.5-mlp2x-336px-pretrain-vicuna-7b-v1.5}{llava-v1.5-mlp2x-336px-pretrain-vicuna-7b-v1.5}, which has no prior exposure to the visual instruction tuning data.} corresponds to the model after training the projector but before any visual instruction tuning in Stage 2. 
Importantly, this model has not been exposed to the \full VIT dataset prior to the data selection process.
In all experiments, we train the models for one epoch following the official finetuning hyperparameters using LoRA. 
More details on computation, including hardware specifications and runtime, are in Appendix~\S\ref{subsec:app_compute}.

\PAR{Target tasks.} 
We evaluate \methodname across diverse multimodal benchmarks (details are in Appendix~\S\ref{sec:exp_details}, Tab.~\ref{tab:datasets}) that test different capabilities of vision-language models:
1) Multiple-choice understanding: MMBench~\citep{mmbench} and MME~\citep{mme}
\footnote{For MME, we focus on its perception section following~\cite{coincide}, which evaluates vision capabilities.}
2) Visual question answering: VQAv2~\citep{vqav2}, GQA~\citep{gqa}, and VizWiz~\citep{vizwiz};
3) Text understanding in images: TextVQA~\citep{textvqa};
4) Scientific reasoning: ScienceQA~\citep{sqa};
5) Open-ended generation: LLaVA-W Bench~\citep{llava};
6) Factual consistency: POPE~\citep{pope}.

\PAR{Baselines.} We compare our \methodname against several baselines, including random selection, CLIP-Score~\cite{clip} for measuring image-text alignment, EL2N~\cite{el2n} based on embedding L2 norms, and Perplexity~\cite{perplexity} using language model scores. 
We also compare against SemDeDup~\cite{semdedup} for semantic deduplication and D2-Pruning~\citep{d2} for distribution-aware pruning. Additional baselines include Self-Sup~\citep{selfsup} leveraging self-supervised signals, while Self-Filter~\citep{self-filter} and COINCIDE~\citep{coincide} are designed for vision-language data selection. 
We reference \full baseline results from COINCIDE~\cite{coincide}.
Additionally, we compare with representation-based data selection approach (RDS)~\cite{ivison2025large, xia2024less}.

\begin{table*}[t!]
    % \vspace{-10pt}
    \centering
    % %\vspace{-5pt}
    \caption{\textbf{Selection results on \full.} Performance comparison of different data selection approaches when trained on 20\% of the \full dataset. Tasks are ordered by random selection's relative performance. The best and second best results for each benchmark are shown in \textbf{bold} and \underline{underlined}, respectively. Our method \methodname achieves the highest overall \relp (98.6\%), consistently outperforming existing approaches including COINCIDE~\cite{coincide} (97.4\%) and D2-Pruning~\cite{d2} (94.8\%).}
    % \vspace{-10pt}
    \resizebox{\textwidth}{!}{
        \begin{tabular}{l |c c c c c c c c c c |c}
             \toprule
             {\textbf{Method}} & {\textbf{MME}} & {\textbf{SQA-I}} & {\textbf{POPE}} & {\textbf{VQAv2}} & {\textbf{LLaVA-W}} & {\textbf{TextVQA}} & \multicolumn{2}{c}{\textbf{MMBench}} & {\textbf{GQA}} & {\textbf{VizWiz}} & {\textbf{\relp (\%)}}\\
             & & & & & {\textbf{Bench}} & & {\textbf{en}} & {\textbf{cn}} & & & \\
             \midrule
             Full & 1476.9 & 68.4 & 86.4 & 79.1 & 67.9 & 58.2 & 66.1 & 58.9 & 63.0 & 47.8 & 100\\
             \cmidrule{0-11}
              Random &
             1483.0 & 68.5 & 84.7 & 75.7 & 65.0 & 55.3 & 62.2 & 54.8 & 58.9 & 44.3 & 95.8\\
            CLIP-Score~\cite{clip} &
             1331.6 & 65.0 & 85.3 & 73.4 & 66.2 & 54.7 & 55.2 & 52.0 & 51.4 & 43.0 & 91.2\\
              EL2N~\cite{el2n} &
             1439.5 & 65.5 & 84.3 & 76.2 & 64.9 & 53.0 & 53.2 & 47.4 & 58.7 & 43.7 & 92.0\\
              Perplexity~\cite{perplexity} &
             1341.4 & 65.1 & 82.6 & 75.8 & \underline{68.3} & 52.8 & 52.0 & 45.8 & 57.0 & 47.8 & 91.6\\
                SemDeDup~\cite{semdedup} &
             1376.9 & 65.8 & 84.7 & 74.2 & \textbf{70.0} & \underline{55.5} & 52.2 & 48.5 & 54.5 & 46.9 & 92.6\\
            D2-Pruning~\cite{d2} &
             1391.2 & \underline{69.3} & 85.7 & 73.0 & 63.9 & 51.8 & \textbf{65.7} & \textbf{57.6} & 58.4 & 41.9 & 94.8\\
            Self-Sup~\cite{selfsup} &
             1335.9 & 67.8 & 83.5 & 74.9 & 63.3 & 49.3 & 61.4 & 53.8 & 59.5 & 46.0 & 93.4\\
                Self-Filter~\cite{self-filter} &
             1306.2 & 61.4 & 83.8 & 73.7 & 64.9 & 52.9 & 48.8 & 45.3 & 58.3 & \textbf{53.2} & 90.9\\
                COINCIDE~\cite{coincide} &
             \textbf{1495.6} & 69.2 & 86.1 & \textbf{76.5} & 67.3 & \textbf{55.6} & \underline{63.1} & 54.5 & \underline{59.8} & 46.8 & \underline{97.4}\\ 
             % \midrule
             RDS~\cite{ivison2025large, xia2024less} & 1393.8 & 68.0 & \underline{86.3} & 75.1 & 63.7 & 54.9 & 61.2 & 52.7 & 57.9 & 48.6 & 95.2\\ 
             \midrule
             \rowcolor[HTML]{E0EFF2} \textbf{\methodname (ours)} & \underline{1485.7} & \textbf{70.8} & \textbf{87.5} & \underline{76.3} & 66.1 & \textbf{55.6} & \underline{63.1} & \underline{55.8} & \textbf{60.7} & \underline{50.1} & \textbf{98.6}\\
             % \textbf{\methodname (ours)} & 76.7 & 61.1 & 50.5 & 71.2 & 55.4 & 86.3 & 1477.3 & 62.4 & 56.3 & 66.5 & 98.7\\ \xindi{maybe appendix, or not}
             \bottomrule
        \end{tabular}
    }
        \vspace{-5pt}
\label{tab:main}
\end{table*}

\subsection{Main results}
\label{subsec:baseline}
%\vspace{-5pt}
\PAR{\full selection.}
As shown in Tab.~\ref{tab:main}, \methodname achieves the best overall performance with 98.6\% \relp on \full, outperforming all baselines with \mini, 20\% of the training data. Remarkably, we achieve comparable or better performance than full dataset training on several tasks: SQA-I (70.8 vs. 68.4), MME (1485.7 vs. 1476.9) and POPE (87.5 vs. 86.4). While COINCIDE achieves strong performance (97.4\% \relp), it falls short of \methodname on key tasks. Approaches like EL2N, Perplexity, SemDeDup achieve only 91-92\% \relp, showing limitations in preserving performance.

\begin{table*}[t!]
    % %\vspace{-10pt}
    \centering
    % %\vspace{-5pt}
        \caption{\textbf{Selection results on \visionflan and \cambrian.} Performance comparison of different data selection approaches when trained on 20\% of the \visionflan~\cite{visionflan} and \cambrian~\cite{tong2024cambrian} datasets. \methodname achieves strong performance (99.8\% and 98.8\% Rel. respectively) while using only 20\% of the training data, significantly outperforming random selection which is one of the strongest baselines, and approaching full performance.}
    \resizebox{\textwidth}{!}{
        \begin{tabular}{c|c|l|c c c c c c c c c c |c}
             \toprule
             {\textbf{Dataset}} & {\textbf{\#Data}} & {\textbf{Method}} & {\textbf{VQAv2}} & {\textbf{GQA}} & {\textbf{VizWiz}} & {\textbf{SQA-I}} & {\textbf{TextVQA}} & {\textbf{POPE}} & {\textbf{MME}} & \multicolumn{2}{c}{\textbf{MMBench}} & {\textbf{LLaVA-W}} & {\textbf{\relp (\%)}}\\
             & & & & & & & & & & {\textbf{en}} & {\textbf{cn}} & {\textbf{Bench}} & \\
             \midrule
             \multirow{3}{*}{\visionflan} 
             & 186k & Full & 68.0 & 49.2 & 41.7 & 60.8 & 50.4 & 83.4 & 1,263.2 & 52.6 & 45.9 & 63.3 & 100.0\\
             \cmidrule{2-14}
             & \multirow{2}{*}{37k} & Random & 64.1 & 45.8 & 37.5 & 58.7 & 45.3 & 82.9 & 1,079.8 & 46.5 & 39.6 & 58.7 & 91.6 \\
            \rowcolor[HTML]{E0EFF2} & & \textbf{ICONS (ours)} & \textbf{67.4} & \textbf{48.8} & \textbf{44.1} & \textbf{60.2} & \textbf{49.9} & \textbf{83.0} & \textbf{1,252.5} & \textbf{51.9} & \textbf{45.4} & \textbf{63.1} & \textbf{99.8}\\
             \midrule
             \multirow{3}{*}{\cambrian} 
             & 7,068k & Full & 80.2 & 62.9 & 58.4 & 75.3 & 60.9 & 86.5 & 1,524.6 & 69.1 & 58.9 & 67.6 & 100.0\\
             \cmidrule{2-14}
             & \multirow{2}{*}{1,414k} & Random & 74.2 & 57.5 & \textbf{61.9} & 71.0 & 57.1 & \textbf{86.4} & 1,465.7 & 63.3 & 49.6 & \textbf{70.4} & 95.4 \\
            \rowcolor[HTML]{E0EFF2} & & \textbf{ICONS (ours)} & \textbf{79.6} & \textbf{62.1} & 60.7 & \textbf{73.9} & \textbf{59.8} & 86.2 & \textbf{1,503.1} & \textbf{67.8} & \textbf{55.8} & 67.0 & \textbf{98.8}\\
             \bottomrule
        \end{tabular}
}
\label{tab:cambrian}
\vspace{-10pt}
\end{table*}
\PAR{\cambrian \& \visionflan selection.}
We provide results on \visionflan~\cite{visionflan} and \cambrian~\cite{tong2024cambrian} in Tab.~\ref{tab:cambrian}. On \visionflan, our method achieves near-full performance (99.8\% \relp) using just 37k samples, significantly outperforming random selection (91.6\%). Similarly, on \cambrian, \methodname maintains strong performance (98.8\% \relp) with 1,414k samples, while random selection achieves 95.4\%. These results demonstrate our approach works effectively on both small and large datasets, consistently preserving model capabilities while drastically reducing training data.

\PAR{Comparisons with representation-based data selection.} 
We compare against \textbf{RDS} (Representation-based Data Selection)~\cite{ivison2025large, xia2024less}, a strong baseline in language-only data selection. 
RDS computes training-validation similarity using final-layer representation of the last token in each sequence instead of gradients.
For a fair comparison, we use the same influence matrix formulation (Eqn.~\ref{eqn:influence_matrix}) and apply majority voting.
Our method consistently outperforms RDS across all tasks, particularly those requiring perceptual grounding -- e.g., higher scores on GQA (60.7 vs. 57.9), SQA-I (70.8 vs. 68.0), and MME (1485.7 vs. 1393.8).
While RDS is effective for text-only data (e.g. TULU-2/3~\cite{ivison2023camels, lambert2024t}), vision-language tasks demand alignment between modalities, where representation similarity is limited as it only reflects current embedding proximity, whereas gradient-based approaches directly estimate each sample's contribution to validation loss, better capturing cross-modal training dynamics. We further provide qualitative comparisons in Appendix~\S\ref{sec:visualization}.

\subsection{Analysis of aggregation strategies}
\label{subsec:ablation}
%\vspace{-10pt}
\PAR{Ablations.}
As introduced in \S\ref{subsec:generalist}, we explore multiple strategies for aggregating task-specific influence rankings into a single compact subset.
We compare our majority voting approach (\textbf{Vote}) with score-based methods (\textbf{Merge}, \textbf{Max}, and normalized variants (\textbf{Merge-SumNorm}, \textbf{Merge-GausNorm})) and rank-based methods (\textbf{Round Robin}, \textbf{MinRank}). Our approach outperforms all alternatives (Tab.~\ref{tab:aggregation-simplified-results}; full results in Appendix Tab.~\ref{tab:aggregation-full-results}), achieving 98.6\% Rel. Our majority voting effectively identifies consistently influential examples without requiring score calibration across tasks.

\begin{table}[t!]
    \centering
    % %\vspace{-10pt}
    \caption{\textbf{Comparison of aggregation approaches.} Relative performance of different influence aggregation methods when selecting 20\% of the LLAVA-665K dataset on the ten tasks from Tables~\ref{tab:main} and~\ref{tab:cambrian}.
    Our \textbf{Vote} achieves the best performance (98.6\% Rel.), outperforming score-based, their normalized variants, and rank-based baselines. See Appendix Tab.~\ref{tab:aggregation-full-results} for per-task results.
    }
    \resizebox{\linewidth}{!}{
    \begin{tabular}{l|cc|cc}
    \toprule
     \textbf{Type} & \multicolumn{2}{c}{\textit{Score-based}} & \multicolumn{2}{c}{\textit{Normalized Score-based}} \\
    \midrule
    Method & Merge & Max & Merge-GausNorm & Merge-SumNorm \\
    \rowcolor[HTML]{E0EFF2} Rel. (\%) & 96.4 & 96.1 & 96.8 & 95.3 \\
    \midrule
     \textbf{Type} & \multicolumn{2}{c}{\textit{Rank-based}} & \multicolumn{2}{c}{\textit{Vote-based}} \\
    \midrule
    Method & Round Robin & MinRank & \multicolumn{2}{c}{\textbf{ICONS (ours)}} \\
    \rowcolor[HTML]{E0EFF2} Rel. (\%) & 96.7 & \underline{97.1} & \multicolumn{2}{c}{\textbf{98.6}} \\
    \bottomrule
    \end{tabular}
    }
    \vspace{-5pt}
\label{tab:aggregation-simplified-results}
\end{table}
\PAR{Limitations of score-based aggregation.} Score-based methods like \textbf{Merge} (Eqn.~\ref{eqn:merge}) and \textbf{Max} (Eqn.~\ref{eqn:max}) assume calibrated influence scores across tasks, which is rarely the case. We observe substantial variation in influence score distributions across tasks with standard deviations spanning from $8.15 \times 10^{-3}$ (MME) to $1.26 \times 10^{-3}$ (VQAv2), indicating that influence scores for MME are much more spread out, while those for VQAv2 are tightly concentrated. Similarly, mean influence scores vary in both magnitude and sign: MME has a relatively high positive mean ($1.68 \times 10^{-3}$), while tasks like POPE ($-2.83 \times 10^{-4}$) and GQA ($-8.89 \times 10^{-5}$) have negative means.
These patterns show certain tasks have wider influence scores distributions, making a sample helpful for one task but neutral or harmful for another. Aggregating raw scores biases selection toward tasks with higher variance or skewed means. To address this calibration issue, we experiment with normalization:
\textbf{Merge-SumNorm}, which rescales each task's influence scores by dividing them by a task-specific normalization factor (i.e., sum), and
\textbf{Merge-GausNorm}, which normalizes the scores using task-wise mean and standard deviation before averaging:
\vspace{-5pt}
\begin{equation*}
    I_{\text{Merge-SumNorm}}(\vz_i) = \sum_{k=1}^K \frac{I_k(\vz_i)}{\sum_{j} I_k(\vz_j)} 
\end{equation*}
\begin{equation*}
     I_{\text{Merge-GausNorm}}(\vz_i) = \sum_{k=1}^K \frac{I_k(\vz_i) - \mu_k}{\sigma_k}
\end{equation*}
However, as shown in Tab.~\ref{tab:aggregation-simplified-results}, both methods still underperform compared to our voting-based strategy, reinforcing the limitations of relying on score magnitudes directly.

\paragraph{Limitations of rank-based aggregation.}
Rank-based methods sidestep the challenge of comparing raw influence scores by focusing on within-task ranking.
\textbf{Round Robin} selects samples by cycling through each task and picking the highest-scoring remaining sample for that task, ensuring balanced coverage. 
\textbf{MinRank} (Eqn.~\ref{eqn:minrank}) selects samples that have the best minimum rank across all tasks, prioritizing examples that perform exceptionally well in at least one task regardless of their performance in others.
Although these methods ensure balanced coverage across tasks, they can overfit to outlier tasks.
This is particularly evident with LLaVA-W Bench~\cite{llava}, which is an outlier in its influence ranking: both Round Robin and MinRank achieve relatively high scores on it (e.g. MinRank: 68.4) but at the cost of lower performance on all other tasks (Tab.~\ref{tab:aggregation-simplified-results}). 
Rank-based selection can trade off overall efficacy on the mainstream tasks.
In contrast, our \textbf{Vote} approach focuses on cross-task consensus rather than forcing equal representation, yielding better balance and higher overall performance (98.6\% Rel.), showing the importance of identifying broadly influential examples rather than optimizing per-task rankings.

% \vspace{-10pt}
\begin{figure}[t]
    \centering
    \vspace{-5pt}
    \includegraphics[width=\linewidth]{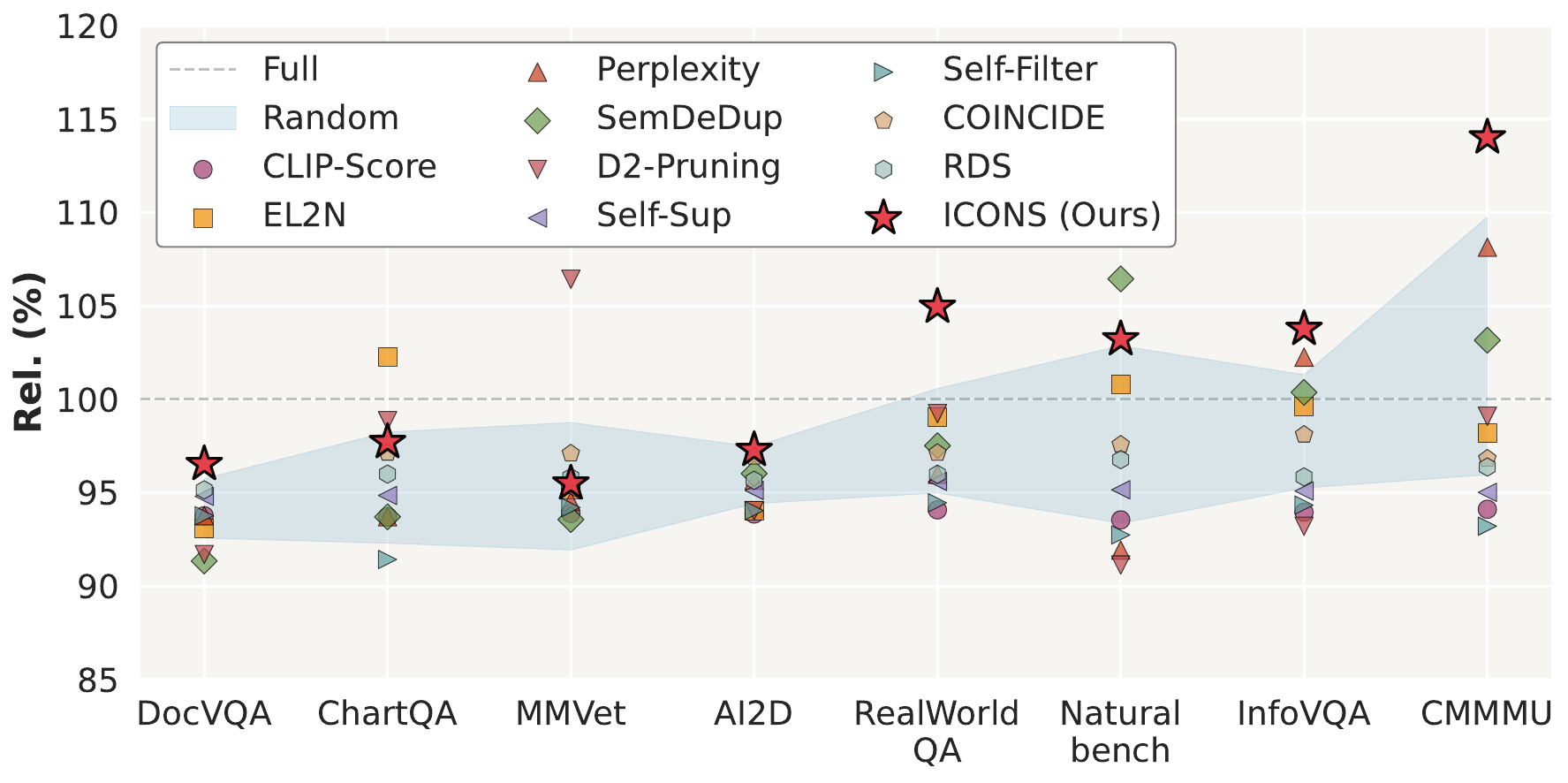}
    \vspace{-17pt}
    \caption{\textbf{Unseen task generalization across methods.} Performance on eight unseen benchmarks using 20\% selected subsets, sorted by the average performance across 5 random selection runs (left: lower, right: higher). Shaded region shows mean and standard deviation across random runs. Our \methodname consistently outperforms all baselines. Full detailed results are in Appendix Tab.~\ref{tab:cross_task}.
    }
    \vspace{-5pt}
    \label{fig:unseen_comparison}
\end{figure}

\subsection{\methodname generalizes to unseen tasks}
\label{subsec:transferability}
%\vspace{-5pt}

Data selected via \methodname demonstrate strong generalization on unseen benchmarks not used during data selection. In Fig.~\ref{fig:unseen_comparison} and Appendix Tab.~\ref{tab:cross_task} we test \mini on diverse tasks including MMVet~\cite{mmvet}, NaturalBench~\cite{naturalbench}, AI2D~\cite{ai2d}, ChartQA~\cite{chartqa}, DocVQA~\cite{docvqa}, InfoVQA~\cite{infographicvqa},  RealWorldQA~\cite{xai2024grok} and CMMMU~\cite{cmmmu}. 
Notably, \mini achieves substantial improvements on challenging benchmarks (MMVet: 95.5\%, AI2D: 97.3\%) and even exceeds full dataset performance on harder tasks (CMMMU: 114.0\%, RealWorldQA: 104.4\%, InfoVQA: 103.8\%, NaturalBench: 103.2\%), demonstrating superior generalization to entirely unseen task formats and domains despite these tasks not being included in the selection process.
Importantly, \mini outperforms random selection across all benchmarks. This suggests that our selection approach successfully captures fundamental visual-language understanding capabilities that transfer well across different task formats and domains. We further provide the cross-architecture generalization results in Appendix \S\ref{sec:add_analysis}.

\subsection{\methodname outperforms baselines across ratios and exceeds full-data training at 60\%} 

\label{subsec:ratio}

\begin{figure}[t]
    \centering
    % \vspace{-10pt}
    \includegraphics[width=\linewidth]{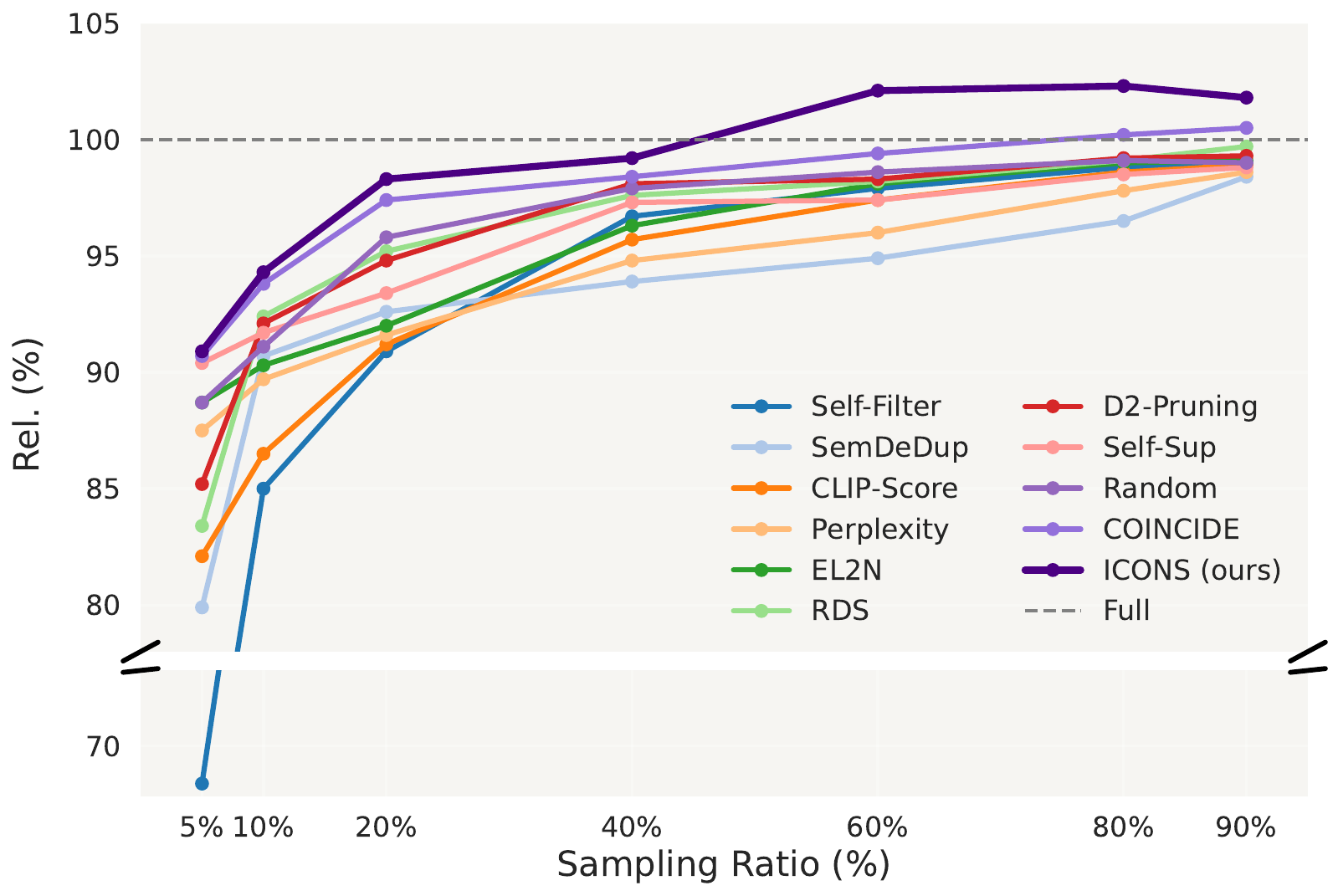}
    \vspace{-20pt}
    \caption{\textbf{Different selection ratios.} \methodname consistently outperforms all baselines across different selection ratios and remarkably exceeds 102\% at a 60\% selection ratio.
    }
    \vspace{-20pt}
    \label{fig:selection_ratio}
\end{figure}

To understand how \methodname scales, we evaluate it across different selection ratios, progressively scaling the subset size from 5\% to 90\% of \full.
As shown in Fig.~\ref{fig:selection_ratio}, our results reveal several key patterns: First, \methodname shows particularly strong performance in the low-selection regime (5-20\%), where identifying the most influential samples is crucial. 
Second, as the selection ratio increases, the performance gap between different methods gradually narrows. This convergence pattern is expected, as larger sample sizes naturally capture more of the dataset's diversity and information. 
Despite this convergence trend, \methodname consistently outperforms all baselines across all selection ratios. 
Remarkably, it even surpasses full dataset performance at the 60\% ratio, achieving over 102\% relative score. One hypothesis is that \methodname can also effectively filter out potentially harmful or noisy training samples that might negatively impact model training, thereby surpassing the full training performance.

\subsection{How does voting change with selection ratio?}
\label{app:vote_evolution}

We analyze vote distribution across selection ratios to understand task consensus patterns (Tab.~\ref{tab:vote_stats}, Appendix Fig.~\ref{fig:vote_evolution}). At 5\% selection, 72.1\% of samples receive zero votes, indicating highly task-specific selection. As the ratio increases to 50\% and 95\%, the distribution shifts rightward with median votes rising to 4 and 5-7 respectively, and zero-vote samples decreasing to 8.3\% and 1.2\%. This rightward shift demonstrates that task consensus emerges as selection criteria relax, revealing samples with broader multi-task utility, while persistent zero-vote samples indicate universally low-quality or redundant data. These patterns validate our voting-based aggregation approach, which prioritizes samples with higher multi-task agreement. We further visualize zero-vote and one-vote samples in \S\ref{subsec:zero}.

\begin{table}[t!]
\centering
\small
% \vspace{-10pt}
\caption{\textbf{Vote distribution across selection ratios.} Vote distributions shift from task-specific selection at low ratios to broader consensus at high ratios.}
\vspace{-7pt}
\label{tab:vote_stats}
\resizebox{\linewidth}{!}{%
\begin{tabular}{l|ccc|c|c}
\toprule
\textbf{Ratio} & Mean (\textcolor{gray}{$\pm$Std}) & Median & Max Votes & Zero-Vote & Threshold \\
\midrule
5\%   & 0.50 \textcolor{gray}{$\pm$0.98} & 0 & 7 & 72.1\% & 3 votes \\
20\%  & 2.00 \textcolor{gray}{$\pm$1.56} & 2 & 7 & 28.4\% & 5 votes \\
50\%  & 5.00 \textcolor{gray}{$\pm$2.21} & 5 & 7 & 8.3\% & 6 votes \\
90\%  & 9.00 \textcolor{gray}{$\pm$1.84} & 9 & 10 & 1.2\% & 9 votes \\
95\%  & 9.50 \textcolor{gray}{$\pm$1.42} & 10 & 10 & 0.8\% & 10 votes \\
\bottomrule
\end{tabular}
}
\vspace{-10pt}
\end{table}

\subsection{What do low-vote training samples look like?}
\label{subsec:zero}

To understand what samples are filtered out, we examine zero-vote and 1-vote samples at a 90\% selection ratio for \full, where the majority of samples receive votes, making the remaining low-vote samples particularly indicative of low-quality data. Zero-vote samples do not rank in the top percentile for any task, while 1-vote samples are selected by only a single task-specific specialist. As shown in Appendix Fig.~\ref{fig:low_vote_samples}, these low-vote samples exhibit characteristics that explain their low selection frequency: (1) subjective questions lacking objective ground truth (e.g., ``Is the woman pretty?''), (2) trivial questions providing minimal visual learning signal (e.g., ``Is the door open?'' with answer ``Yes''), and (3) redundant questions about the same image content. By filtering out such potentially harmful or redundant examples, \methodname focuses training on more informative and generalizable samples, and thus exceeds full-dataset performance.

%\vspace{-5pt}
\section{Conclusion}
%\vspace{-5pt}
We introduce \methodname, an influence consensus-based approach for visual instruction tuning data selection. It leverages gradient-based influence estimation and aggregates task-level influence scores through majority voting to select training samples broadly beneficial across tasks.
\methodname addresses limitations of prior approaches by avoiding assumptions about score comparability across tasks and reducing sensitivity to outlier task rankings that bias selection. %(score-based and rank-based)
\methodname builds compact, high-impact datasets achieving 98.6\% of full performance using only 20\% of \full while generalizing well to unseen tasks and architectures.
Beyond data selection, it provides a principled way to reason about data influence in multitask mixtures.
We release \mini, \cambrianmini, and \visionflanmini as compact 20\% subsets maintaining strong performance, hoping to inspire further exploration into data-efficient vision-language models.

% \newpage
\PAR{Limitations.}
Our approach's main limitation is computational expense: computing gradients for large datasets is costly (Appendix \S\ref{subsec:app_compute}), potentially constraining applicability to extremely large-scale data. To support the research community under resource constraints, we release the \mini dataset for iteration and model development.

\PAR{Broader impact.}
While our work does not directly imply negative impacts, it may indirectly propagate biases present in the original datasets. 
Positively, it enables more efficient and sustainable model development by reducing data redundancy and computational cost while maintaining performance. Continued discussion on these aspects remains critical for data-efficient vision-language models.

\small
{
\PAR{Acknowledgments.} This material is based upon work supported by the National Science Foundation under Grant No. 2107048 and No.2112562. Any opinions, findings, and conclusions, or recommendations expressed in this material are those of the author(s) and do not necessarily reflect the views of the National Science Foundation.
This work is also supported by the Singapore National Research Foundation and the National AI Group in the Singapore Ministry of Digital Development and Information under the AI Visiting Professorship Programme (award number AIVP-2024-001).
We thank many people for their helpful discussion and feedback, listed in alphabetical order by last name: Allison Chen, Hee Seung Hwang, Hamish Ivison, Carlos E. Jimenez, Polina Kirichenko, Jaewoo Lee, Zhiqiu Lin, Tiffany Ling, Shengbang Tong, Ethan Tseng, Esin Tureci, Justin Wang, Zirui Wang.
}

\newpage

\newpage
{
    \small
    \bibliographystyle{unsrt}
    \bibliography{main}
}

\newpage
\appendix
\appendixpage

\hypertarget{toc}{}
\startcontents[sections]
\printcontents[sections]{l}{1}{\setcounter{tocdepth}{2}}
% \tableofcontents

\newpage

\pagestyle{fancy}
\renewcommand{\headrulewidth}{0pt}
\fancyhead{}
\fancyfoot[L]{\hyperlink{toc}{Back to Table of Contents}}
\fancyfoot[R]{\hyperlink{abstract}{Back to the First Page}}

\section{Computational complexity}
\label{subsec:app_compute}
\subsection{Complexity analysis}
Computing gradient-based influence requires a non-trivial amount of computational resources. In the specialist stage, the complexity scales with both the dataset size $|\mathcal{D}|$ and the gradient dimension $d$. This stage consists of three steps. 
First, the warm-up training has a complexity of $\mathcal{O}(|\mathcal{D}_{\text{warmup}}| )$. 
Second, the gradient computation stage has a computational complexity of $\mathcal{O}(|\mathcal{D}| + |\mathcal{D}_{\text{val}}|)$ for forward and backward passes, with storage requirements of $\mathcal{O}(|\mathcal{D}| \cdot d + |\mathcal{D}_{\text{val}}| \cdot d)$ for the gradients. 
Third (and finally), the influence matrix computation requires $\mathcal{O}(|\mathcal{D}| \cdot |\mathcal{D}_{\text{val}}| \cdot d')$ compute cost, where $d'$ is the reduced dimension after projection. 
The generalist stage, focusing on influence consensus across tasks, has lower computational requirements. It begins with threshold computation, requiring $\mathcal{O}(K \cdot |\mathcal{D}| \log |\mathcal{D}|)$ operations for sorting across $K$ tasks. 
The voting process then takes $\mathcal{O}(K \cdot |\mathcal{D}|)$ compute, followed by a final selection step with complexity $\mathcal{O}(|\mathcal{D}| \log |\mathcal{D}|)$ for sorting the aggregated votes. Storage requirements for this stage are minimal, primarily for the final selected subset.

\subsection{Resource requirements}
In practice, for LLaVA-665K training data, the warmup training phase requires 0.75 hours using eight L40 GPUs. We parallelize the gradient computation across 100 A6000 GPUs, taking approximately one hour and requiring 103GB of total storage for the gradients. The influence consensus stage is notably efficient, completing in less than a minute on a single L40 GPU. While these computational demands are substantial, they represent front-loaded, one-time costs that can be used across multiple target tasks and model iterations. This makes our method extendable for new tasks, as the expensive training data gradient computations only need to be performed once.

\begin{table*}[t!]
    \centering
    % %\vspace{-10pt}
    \caption{\textbf{Detailed aggregation approaches comparison for Tab.~\ref{tab:aggregation-simplified-results}.} Our proposed aggregation approach (\textbf{Vote}) consistently achieves the best overall performance (98.6\% Rel.), outperforming both score-based (\textbf{Merge}, \textbf{Max}), their normalized variants (\textbf{Merge-GausNorm}, \textbf{Merge-SumNorm}) and rank-based (\textbf{Round Robin}, \textbf{MinRank}) baselines when selecting 20\% of the LLAVA-665K dataset.
    }
    \resizebox{\textwidth}{!}{
    \begin{tabular}{l|cccccccccc|c}
    \toprule
    \textbf{Aggregation} & \textbf{VQAv2} & \textbf{GQA} & \textbf{VizWiz} & \textbf{SQA-I} & \textbf{TextVQA} & \textbf{POPE} & \textbf{MME} & \textbf{MMB (en)} & \textbf{MMB (cn)} & \textbf{LLaVA-W} & \textbf{Rel. (\%)} \\
    \midrule
    % Direct merge & 76.1 & 59.4 & 46.1 & 68.7 & 54.1 & 85.1 & 1419.2 & 61.9 & 50.3 & 65.2 & 94.7 \\ 
    Full & 79.1 & 63.0 & 47.8 & 68.4 & 58.2 & 86.4 & 1476.9 & 66.1 & 58.9 & 67.9 & 100\\ \midrule
    Merge & \underline{75.7} & 59.6 & 47.9 & 65.5 & \underline{55.5} & 86.0 & 1422.1 & 59.0 & 51.0 & 66.2 & 96.4 \\
    Max & 75.2 & 59.8 & 48.1 & 66.2 & \underline{55.5} & 85.5 & 1470.7 & 58.3 & 51.8 & 66.2 & 96.1 \\
    Merge-GausNorm & 75.1 & \underline{60.1} & 46.4 & 69.8 & 54.5 & 85.6 & \underline{1482.6} & 58.9 & \underline{52.5} & 66.3 & 96.8 \\
    Merge-SumNorm & 75.5 & 59.1 & \textbf{51.7} & 68.7&  43.5 & \underline{87.1} & 1478.3 & 59.5 & 50.9 &\textbf{69.8} & 95.3\\
    % Rank & 74.9 & 58.6 & 40.5 & \underline{69.8} & 55.2 & 85.6 & \textbf{1490.0} & \underline{59.0} & 50.8 & \textbf{66.4} & 95.9 \\
    Round Robin & 75.4 & 59.1 &48.3 &\underline{70.6} & 55.2& 86.6& 1474.5 & \underline{61.6} &51.5&66.9 & 96.7\\
    MinRank & 75.2 & 59.0 & 49.7 & 70.4 & 55.1 & 86.9 &1456.3& 61.1& 52.4& \underline{68.4} & \underline{97.1}\\
    % Continuous Majority (w 20\% cap)  & 75.3 & 59.8 & 48.1 & 70.1 & 44.2 & 86.8 &1503.1 & 62.9 & 53.0 & 64.8 & 95.2\\
    \textbf{Vote (ours)} & \textbf{76.3} & \textbf{60.7} & \underline{50.1} & \textbf{70.8} & \textbf{55.6} & \textbf{87.5} & \textbf{1485.7} & \textbf{63.1} & \textbf{55.8} & 66.1 & \textbf{98.6} \\
    \bottomrule
    \end{tabular}
    }
    %\vspace{-5pt}
    \label{tab:aggregation-full-results}
    \end{table*}

\subsection{Discussion on cost-benefit justification}
Although gradient-based data selection is computationally intensive, we argue that the initial cost is justified by three key considerations. First, the computational expense is largely a one-time investment: once gradients are computed, they can be stored in our gradient datastore and reused across multiple model iterations, target tasks, and diverse downstream applications. This reusability becomes especially valuable as the number of target datasets grows, because each new target dataset can leverage existing gradient computations, making the selection increasingly efficient at scale.

Second, our empirical results demonstrate substantial performance benefits. Training on a strategically chosen 60\% subset of data not only reduces training time but also surpasses the performance obtained by using the full dataset. This improvement underscores how directing more compute resources toward a carefully selected subset can yield higher returns on a fixed set of data.

Lastly, the initial compute-intensive investment in data selection is amortized across future training iterations and future developers. By leveraging the curated, higher-quality dataset, they can substantially reduce training costs.

\begin{table}[t!]
    \centering
    \caption{\textbf{Statistics of Target Tasks.} Our target tasks include diverse benchmarks and answer formats, covering different vision-language capabilities. Task types include Multiple-Choice Questions (\textbf{MCQ}), Visual Question Answering (\textbf{VQA}), and Yes/No Questions (\textbf{Y/N}).}
    \resizebox{\linewidth}{!}{
    \begin{tabular}{c| c c c c c}
        \toprule
        \textbf{Task} & \textbf{MME} & \textbf{POPE} & \textbf{SQA-I} & \multicolumn{2}{c}{\textbf{MMBench}} \\
        \cmidrule(lr){5-6}
        & & & & \textbf{en} & \textbf{cn} \\
        \midrule
        $|\mathcal{D}_{\text{val}}|$ & 986 & 500 & 424 & 1,164 & 1,164 \\ 
        $|\mathcal{D}_{\text{test}}|$ & 2,374 & 8,910 & 4,241 & 1,784 & 1,784 \\
        \rowcolor{Gray} \textbf{Task Type} & Y/N & Y/N & MCQ & MCQ & MCQ \\
        \midrule
        \textbf{Task} & \textbf{VQAv2} & \textbf{GQA} & \textbf{VizWiz} & \textbf{TextVQA} & \textbf{LLaVA-W} \\
        \midrule
        $|\mathcal{D}_{\text{val}}|$ & 1,000 & 398 & 800 & 84 & 84 \\ 
        $|\mathcal{D}_{\text{test}}|$ & 36,807 & 12,578 & 8,000 & 5,000 & 84 \\
        \rowcolor{Gray} \textbf{Task Type} & VQA & VQA & VQA & VQA & VQA \\
        \bottomrule
    \end{tabular}
    }
    \vspace{-10pt}
    \label{tab:datasets}
\end{table}

\section{Additional experiment details \& ablations}
\label{sec:exp_details}

We provide comprehensive experimental details and ablation studies to support our main findings. We first present detailed task descriptions and dataset statistics (Tab.~\ref{tab:datasets}), followed by detailed per-task results for aggregation method comparisons (Tab.~\ref{tab:aggregation-full-results}) and full results for unseen benchmarks (Tab.~\ref{tab:cross_task}). We then ablate key hyperparameters including projection dimension (\S\ref{subsec:projection_dim}) and warm-up ratio (\S\ref{subsec:warmup_ratio}).

\begin{table*}[t!]
    \centering
    \vspace{-10pt}
    \caption{\textbf{Detailed results of unseen task generalization.} Detailed results for Fig.~\ref{fig:unseen_comparison}. Performance comparison on unseen benchmarks when trained on selected 20\% subsets. Notably, we observe improvements on InfoVQA (103.8\%), RealWorldQA (104.4\%), and CMMMU (114.0\%), highlighting strong generalization to unseen tasks.
    }
    \resizebox{\linewidth}{!}{
    \begin{tabular}{l|cccccccc|c}
    \toprule
    & \textbf{AI2D} & \textbf{ChartQA} & \textbf{DocVQA}& \textbf{InfoVQA}
    & \textbf{MMVet} & \textbf{Naturalbench} & \textbf{RealworldQA} & \textbf{CMMMU} & \textbf{Rel. (\%)} \\
    \midrule
    Full & 55.4 & 17.5 & 28.9 & 26.5 & 31.1 & 12.4 & 52.4 & 22.1 & 100.0 \\
    \midrule
    Random & 50.2 & 15.1 & 25.2 & 24.3 & 29.6  & 11.1 & 49.8 & 21.9  & 91.8 \\
    CLIP-Score & 52.0 & 16.4 & 27.1 & 24.9 & 29.2 & 11.6 & 49.3 & 20.8 & 93.9 \\
    EL2N & 52.1 & 17.9 & 26.9 & 26.4 & 29.6 & 12.5 & 51.9 & 21.7 & 97.8 \\
    Perplexity & 53.0 & 16.4 & 27.1 & 27.1 & 29.4 & 11.4 & 50.3 & 23.9 & 97.0\\
    SemDeDup & 53.2 & 16.4 & 26.4 & 26.6 & 29.1 & 13.2 & 51.1 & 22.8 & 97.8 \\
    D2-Pruning & 52.1 & 17.3 & 26.5 & 24.7 & 33.1 & 11.3 & 52.0 & 21.9 & 96.7 \\
    Self-Sup & 52.7 & 16.6 & 27.4 & 25.2 & 29.6 & 11.8 & 50.1 & 21.0 & 95.0 \\
    Self-Filter & 52.1 & 16.0 & 27.1 & 25.0 & 29.3 & 11.5 & 49.5 & 20.6 & 94.0 \\
    COINCIDE & 53.7 & 17.0 & 27.9 & 26.0 & 30.2 & 12.1 & 50.9 & 21.4 & 97.2 \\
    RDS & 53.0 & 16.8 & 27.5 & 25.4 & 29.8 & 12.0 & 50.3 & 21.3 & 95.6 \\    
    \midrule
    \mini & 53.9 & 17.1 & 27.9 & 27.5 & 29.7 & 12.8 & 55.0 & 25.2 & 101.6 \\
    \midrule
    \rowcolor{Gray} Per-task Rel. (\%) & 97.3 & 97.7 & 96.5 & 103.8 & 95.5 & 103.2 & 104.4 & 114.0 & - \\
    \bottomrule
    \end{tabular}
    }
    % \vspace{-10pt}
    \label{tab:cross_task}
\end{table*}

\begin{figure*}[h!]
    \centering
    \begin{subfigure}[t]{0.48\linewidth}
        \centering
        \includegraphics[width=\linewidth]{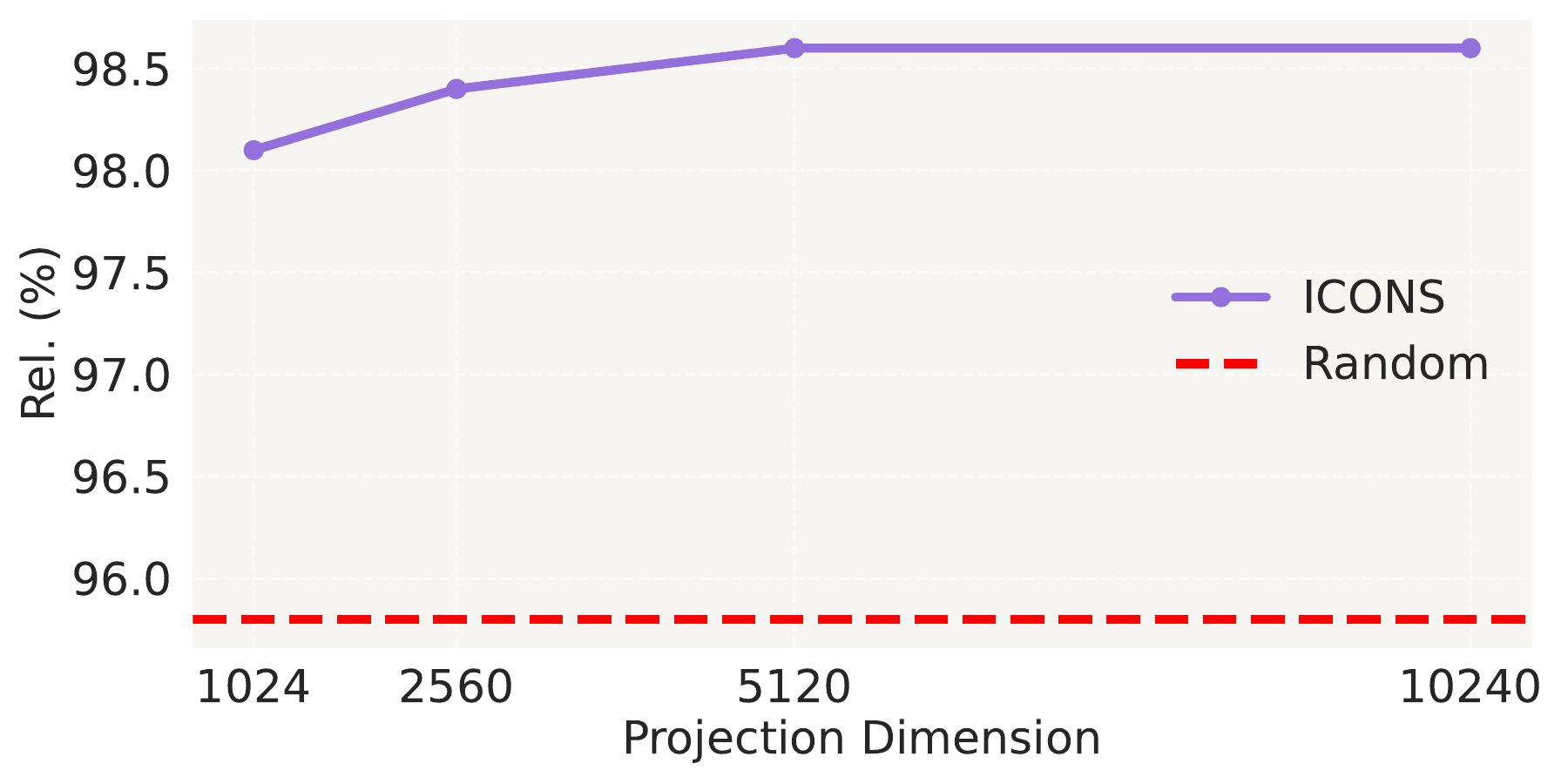}
        \caption{\textbf{Projection Dimension Ablation.} We show the performance of ICONS at different projected dimensions (1024, 2560, 5120, 10240), compared to the random baseline. The performance increases with the projected dimension and reaches a plateau around dimension 5120.}
        \label{fig:projection_dim}
    \end{subfigure}
    \hfill
    \begin{subfigure}[t]{0.48\linewidth}
        \centering
        \includegraphics[width=\linewidth]{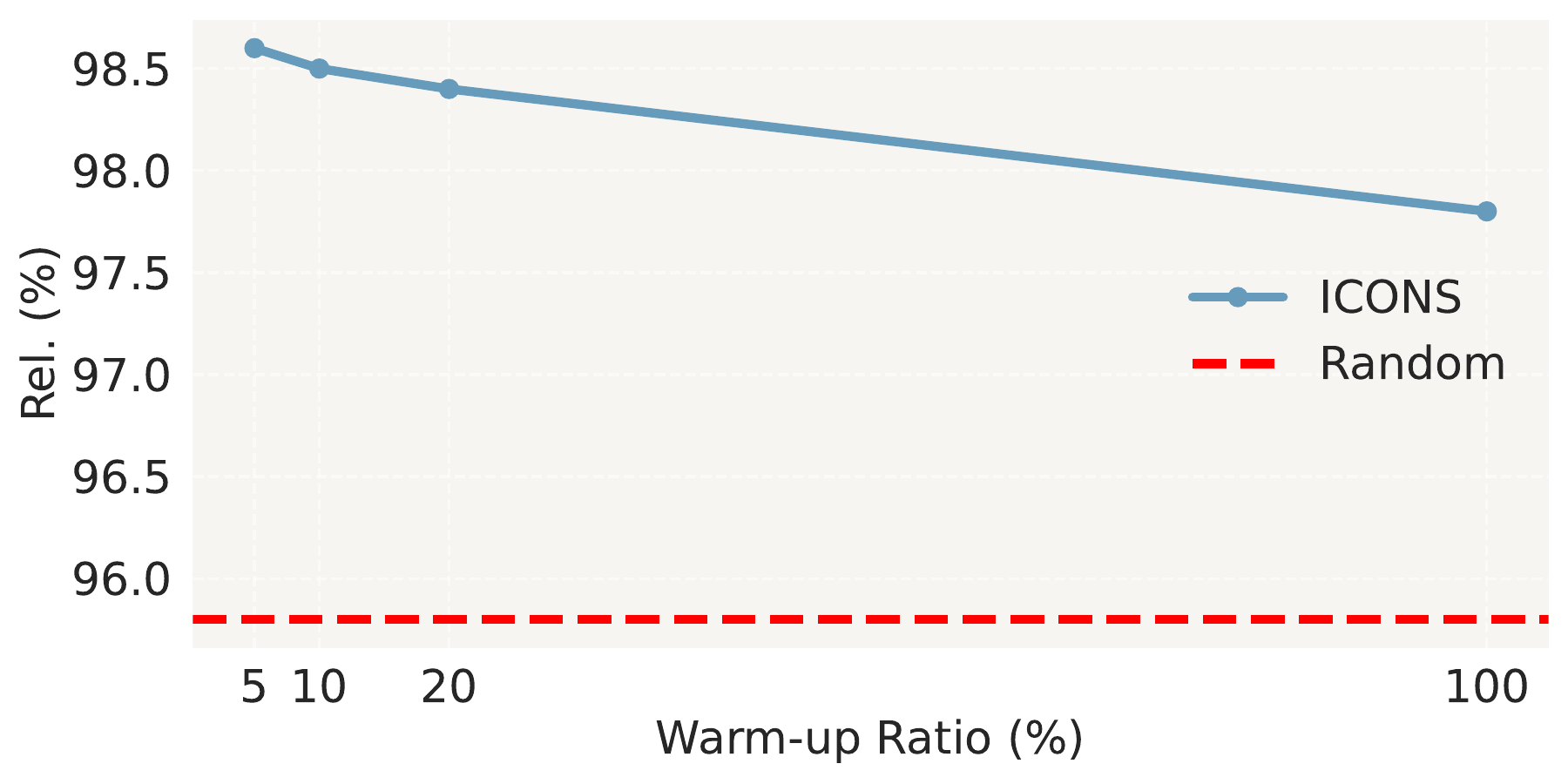}
        \caption{\textbf{Warm-up Ratio Ablation.} The blue line represents ICONS performance across different warm-up ratios (5\%, 10\%, 20\%, and 100\%), while the red dashed line shows the random baseline performance. Results show that smaller warm-up ratios (5-20\%) achieve better performance compared to using the full dataset (100\%).}
        \label{fig:warm_up_ratio}
    \end{subfigure}
    \vspace{-10pt} 
    \caption{Ablation studies on (\textit{left}) projection dimension and (\textit{right}) warm-up ratio.}
    \label{fig:projection_warmup_ablation}
\end{figure*}

\subsection{Additional task details}
Here, we provide further details on the target tasks, as summarized in Tab.~\ref{tab:datasets}. These tasks cover a wide range of multimodal benchmarks commonly used, including Yes/No questions (Y/N), multiple-choice understanding questions (MCQ) and visual question answering (VQA).

\subsection{Projection dimension}
\label{subsec:projection_dim}
We primarily set the projection dimension to 5120, reducing features from 338.7M to 5120 dimensions. The choice of 5120 was empirically validated for its trade-off between effectively capturing gradient representations and maintaining a manageable parameter space. Our \texttt{LLaVA-v1.5-7b-lora} architecture includes a total of 7.4B parameters, with 338.7M parameters being trainable after LoRA adaptation, accounting for approximately 4.58\% of the total parameter count. We further ablate different projection dimensions (1024, 2560, 5120, and 10240), with results provided in Fig.~\ref{fig:projection_dim}.

\subsection{Warm-up Ratio}
\label{subsec:warmup_ratio}
To initiate training, we use 5\% of the total training data. We conducted ablation studies to evaluate the impact of varying warm-up ratios (5\%, 10\%, 20\%, and 100\%) on selection performance, as shown in Fig.~\ref{fig:warm_up_ratio}. Our experiments reveal that increasing the warm-up data size does not lead to performance improvements. Surprisingly, models trained with smaller warm-up ratios (5-20\%) consistently outperform those trained with the full dataset (100\%). Specifically, the 5\% warm-up ratio achieves the best performance at 98.6\%, while using the complete dataset results in a performance drop to 97.8\%. This finding suggests that a small subset of training data is sufficient and even beneficial for model initialization, and potentially gives better signals in the early training stages.

\begin{table*}[t!]
\centering
\caption{\textbf{Single-task Selection (Specialist) vs. Consensus-aware Multi-task Selection (Generalist).} The single-task data selection approach selects 20\% of \full per task, while our consensus-aware multi-task data selection approach selects a total of 20\% data across all tasks.}
\resizebox{\linewidth}{!}{
\renewcommand{\arraystretch}{1.2}
\begin{tabular}{l|cccccccccc}
\toprule
\textbf{Method} & \textbf{VQAv2} & \textbf{GQA} & \textbf{VizWiz} & \textbf{SQA-I} & \textbf{TextVQA} & \textbf{POPE} & \textbf{MME} & \textbf{MMBench (en)} & \textbf{MMBench (cn)} & \textbf{LLaVA-W Bench} \\
\midrule
\textbf{Full} & 79.1 & 63.0 & 47.8 & 68.4 & 58.2 & 86.4 & 1476.9 & 66.1 & 58.9 & 67.9 \\ \midrule
\textbf{Specialist} & 77.1 & 61.1 & 53.1 & 69.8 & 55.7 & 86.6 & 1506.1 & 66.0 & 56.4 & 67.1 \\
\textbf{Generalist} & 76.3 & 60.7 & 50.1 & 70.8 & 55.6 & 87.5 & 1485.7 & 63.1 & 55.8 & 66.1 \\ \midrule
\rowcolor{Gray} \textbf{Delta (\%)} & 1.04 & 0.65 & 5.65 & -1.43 & 0.18 & -1.04 & 1.35 & 4.34 & 1.06 & 1.49 \\
\bottomrule 
\end{tabular}
} 
\vspace{-10pt}
\label{tab:specialist_vs_generalist}
\end{table*}

\section{Additional analysis}
\label{sec:add_analysis}

\begin{figure}[t]
    \centering
    %\vspace{-10pt}
    \includegraphics[width=\linewidth]{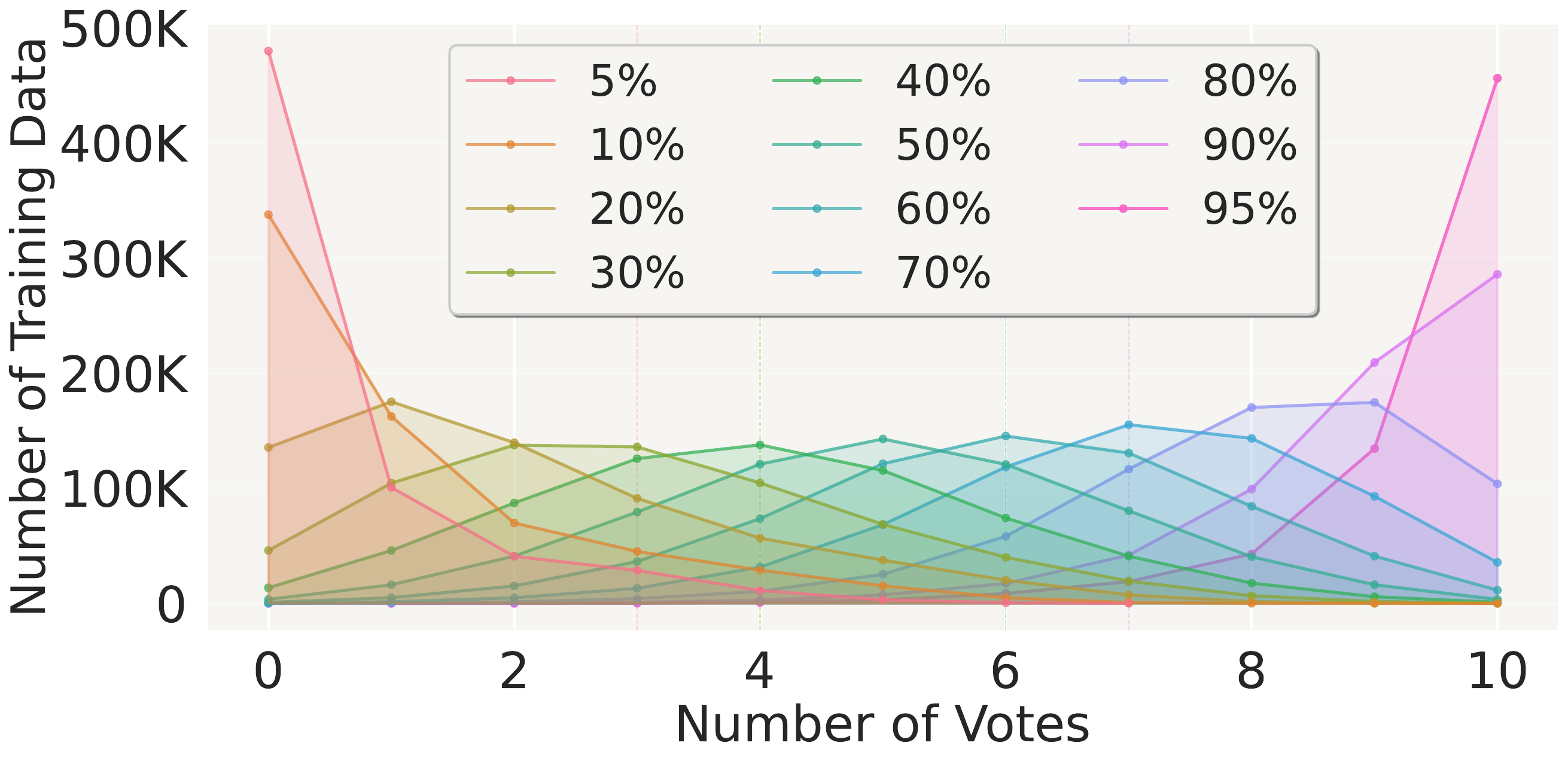}
    %\vspace{-13pt}
    \caption{\textbf{Vote distributions across selection ratios.} 
    Each line shows the vote distribution for a different selection ratio (5\% to 95\%). At lower ratios (e.g., 5\%), most samples receive zero votes, resulting in a left-skewed distribution. As ratios increase (e.g., 50\%, 70\%, 90\%), the distribution shifts rightward and flattens, with more samples receiving votes from multiple tasks. 
    Stricter criteria (lower ratios) identify task-specific samples, while relaxed criteria (higher ratios) reveal broader multi-task agreement. 
    }
    \label{fig:vote_evolution}
\end{figure}

In this section, we provide additional analyses to better understand different aspects of our \methodname. 
We include detailed vote distribution visualizations (Fig.~\ref{fig:vote_evolution}) extending the analysis from \S\ref{app:vote_evolution}.
We analyze the effectiveness of task-specific selections and their aggregation into a generalist subset (\S\ref{subsec:from_specialist_to_generalist}).
We further explore whether incorporating visual dependency information into the selection process affects performance across different types of vision-language tasks (\S\ref{subsec:visual_dependency}). 
We also evaluate the transferability of our selected subset across different model scales (\S\ref{subsec:cross-arch}).
Additionally, we evaluate the consistency of our method across multiple runs (\S\ref{subsec:consistency}), showing its robustness and reliability.

\subsection{From specialist to generalist} 
\label{subsec:from_specialist_to_generalist}
To understand the intermediate task-specific influence matrices we obtained from the specialist stage, we select 20\% of data for each individual task.
The task-specific data (Specialists) achieves comparable or superior performance than full data training (Tab. \ref{tab:specialist_vs_generalist}).
With influence consensus at the generalist stage, we select a 20\% subset with only a 1.33\% average drop across tasks compared to specialist baselines.
This validates our approach: by understanding task-specific influence patterns and building consensus across tasks, we can identify a compact, universal training set that maintains strong performance with significantly less data.

\begin{figure}[t]
    \centering
    \includegraphics[width=\linewidth]{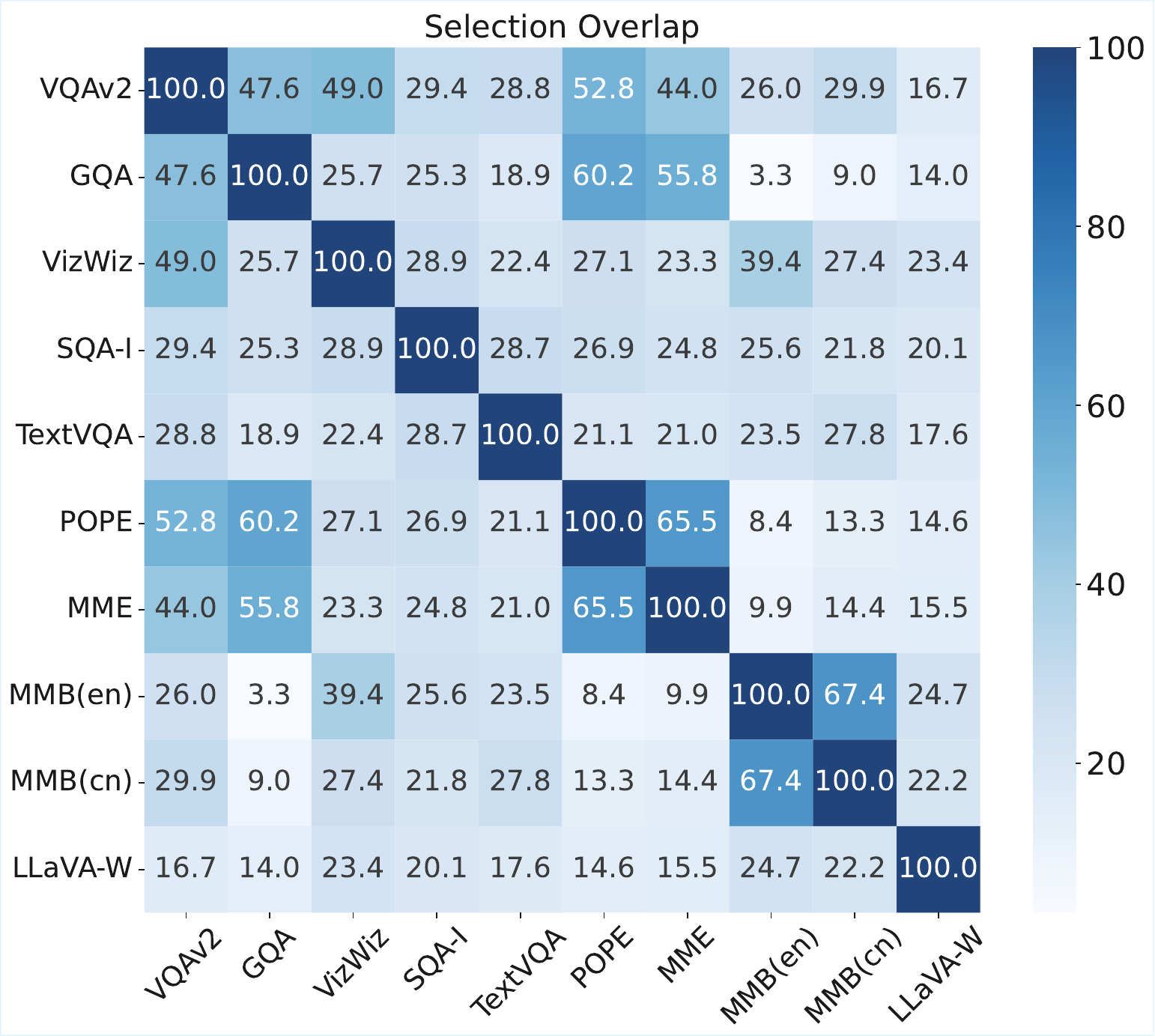}
    \vspace{-15pt}
    \caption{\textbf{Pairwise overlap heatmap between specialists.} 
    The values show overlap percentages between benchmarks' selected samples.
    }
    \label{fig:specialist_overlap_heatmap}
\end{figure}

\begin{figure}[t]
    \centering
    \includegraphics[width=\linewidth]{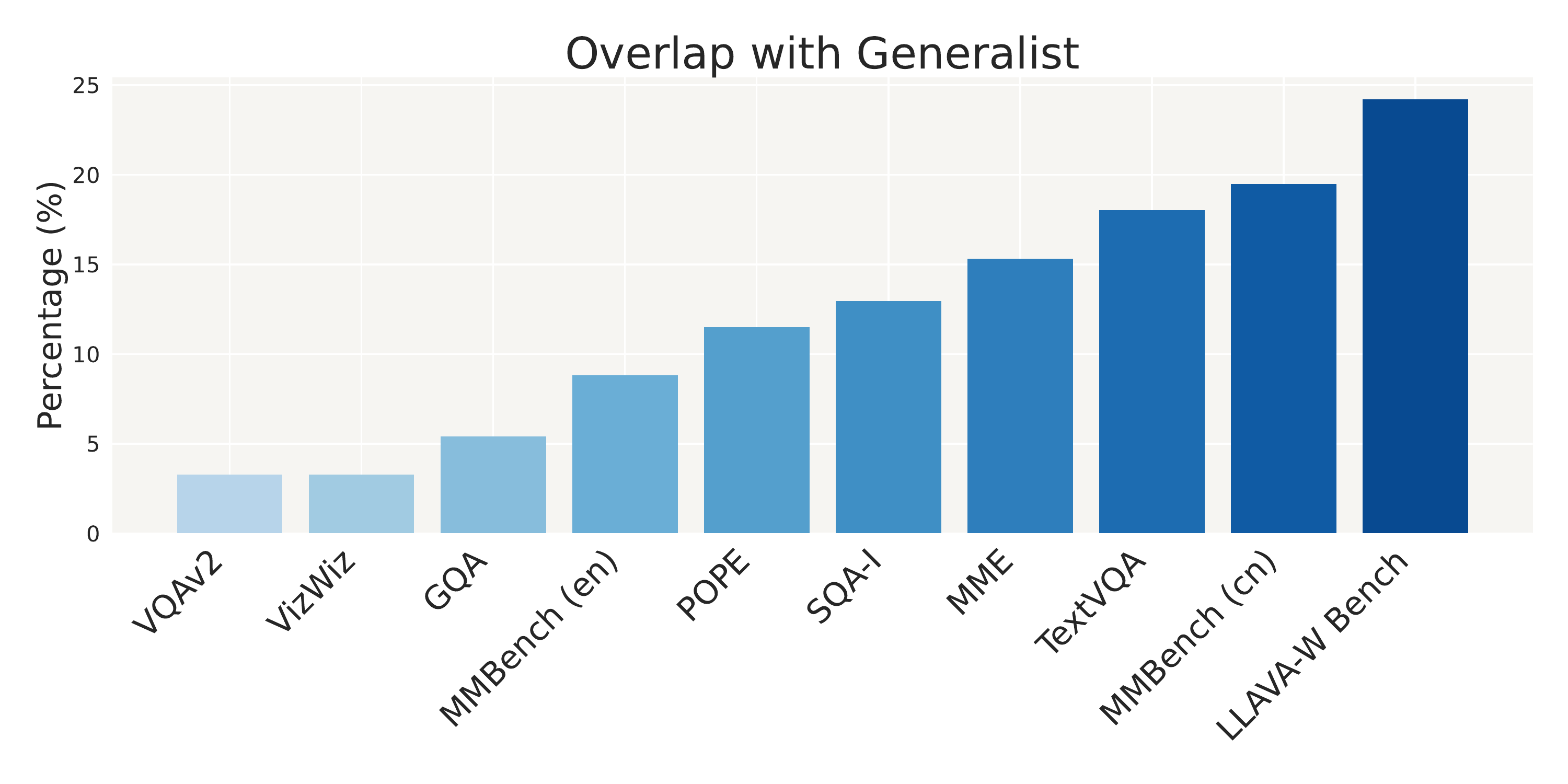}
    \vspace{-20pt}
    \caption{\textbf{Data overlap between specialists and generalist selection.} Overlap varies significantly, from 3.27\% (VQAv2) to 24.21\% (LLAVA-W Bench), reflecting varying alignment between task-specific and consensus selections.
    }
    \label{fig:data_overlap}
\end{figure}

\PAR{Divergent multi-task influence patterns.}
As shown in Fig.~\ref{fig:specialist_overlap_heatmap}, the pairwise overlap heatmap shows notable variation in training data influence across tasks. High overlap -- e.g., VQAv2 and VizWiz (49.0\%) or POPE and GQA (60.2\%), suggests that certain samples are beneficial across similar tasks. However, low overlap, like the 3.3\% between MMBench (en) and GQA, highlights that highly influential samples for one task may have limited impact on others. Even closely related tasks, such as MMBench in different languages (English and Chinese), share 67.4\% of influential samples. 
To understand task-specific influence matrices from the specialist stage, we select the top 20\% samples per task (Specialists).
Overlap with our generalist subset (Fig.~\ref{fig:data_overlap}) varies significantly, from minimal in tasks like VQAv2 (3.27\%) and VizWiz (3.28\%) to substantial agreement in tasks like LLAVA-W Bench~\cite{llava} (24.21\%). 
These findings empirically demonstrate significant overlap in influential samples across tasks and validate our approach: by analyzing task-specific gradient-based influence patterns and building consensus across tasks, we can identify a compact subset that captures broadly useful samples across tasks, yielding strong performance with significantly less data. 

\begin{table*}[t]
\centering
%\vspace{-5pt}
\resizebox{\linewidth}{!}{
\begin{tabular}{l|ccccccccccc}
\toprule
& \textbf{VQAv2} & \textbf{GQA} & \textbf{VizWiz} & \textbf{SQA-I} & \textbf{TextVQA} & \textbf{POPE} & \textbf{MME} & \textbf{MMBench (en)}  & \textbf{MMBench (cn)}  & \textbf{LLaVA-W Bench} \\\midrule
\textbf{Full} & 79.1 & 63.0 & 47.8 & 68.4 & 58.2 & 86.4 & 1476.9 & 66.1 & 58.9 & 67.9 \\ \midrule
\textbf{w/o VDS} & 76.3 & 60.7 & 50.1 & 70.8 & 55.6 & 87.5 & 1485.7 & 63.1 & 55.8 & 66.1 \\
\textbf{w/ VDS} & 75.8 & 60.9 & 50.3 & 69.5 & 54.8 & 86.8 & 1489.3 & 64.3 & 56.3 & 67.9 \\ \midrule
\textbf{Delta (\%)} & -0.66 & +0.33 & +0.40 & -1.84 & -1.44 & -0.80 & +0.24 & +1.90 & +0.90 & +2.72 \\
\bottomrule
\end{tabular}
}
%\vspace{5pt}
\caption{\textbf{Impact of Visual Dependency Score (VDS) on Selection Performance.} Rows show performance without VDS, with VDS, and the performance change (Delta). VDS improves performance on LLaVA-W Bench (+2.72\%), MMBench (en) (+1.90\%), and MMBench (cn) (+0.90\%), but decreases performance on SQA-I (-1.84\%), TextVQA (-1.44\%), and POPE (-0.80\%).}
% %\vspace{-10pt}
\label{tab:vds_ablation}
\end{table*}

\subsection{Visual dependency influence ranking}
\label{subsec:visual_dependency}
Recent work~\cite{tong2024cambrian} has shown that vision-language tasks vary in their reliance on visual information: tasks like MMBench~\cite{mmbench} depend heavily on visual grounding, while others like SQA-I~\cite{sqa} can be handled primarily through language, showing only a 5\% drop in performance when visual input is removed~\cite{tong2024cambrian}. To take visual dependency of training data into account, we further explored gradient-based Visual Dependency Score (\textbf{VDS}). For each data point, we calculate the gradient of the model’s auto-regressive cross-entropy loss with both the original image and a Gaussian noise image $I_{\text{noise}} \sim \mathcal{N}(0, 1)$, keeping the text input constant. This quantifies how much the visual component contributes to model performance. We construct an adapted influence matrix: visual influence matrix $\mathcal{I}_{\texttt{VDS}} \in \mathbb{R}^{|\mathcal{D}| \times |\mathcal{D}_{\text{val}}|}$, which quantifies the visual influence of each training sample $\vz_i$ on each validation sample $\vz'_j$ w.r.t the model’s gradient alignment and visual dependency.
$\mathcal{I}_{\text{VDS}}$ is computed as:
\begin{equation}
    \mathcal{I}_{\texttt{VDS}, ij} = \langle \nabla_{\theta} \mathcal{L}(\vz'_i),\nabla_{\theta}\mathcal{L}(\vx_j,I_j) - \nabla_{\theta}\mathcal{L}(\vx_j,I_{\text{noise}}) \rangle, 
\end{equation}
where $\nabla_{\theta}\mathcal{L}(\vx_j, I_j)$ and $\nabla_{\theta}\mathcal{L}(\vx_j, I_{\text{noise}})$ are the gradients computed with the original and Gaussian noise images, respectively. The visual influence matrix $\mathcal{I}_{\texttt{VDS}}$ provides insights into which training samples have the most influence on the validation samples from a visual perspective. This matrix can be used to further rank and select training data that are most impactful for tasks requiring strong visual grounding, ensuring that the selected subset effectively supports vision-dependent performance. 

Our empirical results demonstrate that VDS-based data selection has varying effectiveness across different tasks (Tab.~\ref{tab:vds_ablation}). It shows substantial improvements on tasks requiring strong visual understanding, such as open-ended generation (LLaVA-W Bench: +2.72\%) and multiple-choice understanding (MMBench-EN: +1.90\%, MMBench-CN: +0.90\%). However, tasks that primarily rely on textual reasoning show decreased performance, including SQA-I (-1.84\%) and TextVQA (-1.44\%). These results align with and extend the findings in Cambrian~\cite{tong2024cambrian}, demonstrating that the effectiveness of VDS corresponds to a task's visual dependency - tasks that maintain performance without visual inputs show limited or negative impact from VDS-based selection, while visually-dependent tasks benefit significantly. This pattern suggests that VDS effectively identifies training samples where visual information plays an important role in training.

\begin{table*}[t!]
\centering
%\vspace{-5pt}
\caption{\textbf{Cross-Architecture Generalization.} Our \mini selected via LLaVA-v1.5-7B model (7B-selected) shows strong cross-architecture transferability, achieving 97.3\% \relp, while the data selected via LLaVA-v1.5-13B model (13B-selected) reaches 98.1\%, showing that our selected subset generalizes well to different architectures. }
\resizebox{\linewidth}{!}{
\begin{tabular}{l|cccccccccc|c}
\toprule
& \textbf{VQAv2} & \textbf{GQA} & \textbf{Vizwiz} & \textbf{SQA-I} & \textbf{TextVQA} & \textbf{POPE} & \textbf{MME} & \textbf{MMBench (en)} & \textbf{MMBench (cn)} & \textbf{LLAVA-W} & \textbf{Rel. (\%)} \\
\midrule
\textbf{Full} & 80.0 & 63.3 & 58.9 & 71.2 & 60.2 & 86.7 & 1541.7 & 68.5 & 61.5 & 69.5 & 100.0 \\
\cmidrule{0-11}
\textbf{Random} & 77.3 & 60.7 & 57.6 & 69.1 & 56.8 & 82.9 & 1517.2 & 63.2 & 56.3 & 67.5 & 95.7 \\
\textbf{7B-selected} & 78.8 & 60.4 & 57.4 & 70.4 & 58.3 & 84.3 & 1527.5 & 64.9 & 59.7 & 68.2 & 97.3 \\
\textbf{13B-selected} & 78.9 & 61.2 & 57.5 & 71.3 & 58.4 & 85.9 & 1535.2 & 66.1 & 59.8 & 68.8 & 98.1 \\
\bottomrule
\end{tabular}
}
%\vspace{-13pt}
\label{tab:13b}
\end{table*}

\subsection{Cross-architecture generalization} 
\label{subsec:cross-arch}
We conduct experiments on cross architecture generalization to evaluate the transferability of our selected data across different model scales. While our subset was initially selected using LLaVA-v1.5-7B as the base model, we investigate whether these same examples remain effective for training larger architectures like LLaVA-v1.5-13B. This tests whether our selection criteria identify universally valuable training examples rather than model-specific patterns. Our results in Tab.~\ref{tab:13b} show cross-architecture generalization, with 13B model trained on 7B-selected subset achieving 98.1\% \relp. Both 7B-selected and 13B-selected subsets outperform random selection (95.7\%), with the 13B-selected option showing particular strength in reasoning tasks like MMBench and POPE. 
This suggests our selected subset captures fundamental visual-language understanding patterns that generalize well across different model architectures.

\begin{figure}[t]
    \centering
    %\vspace{-10pt}
    \includegraphics[width=\linewidth]{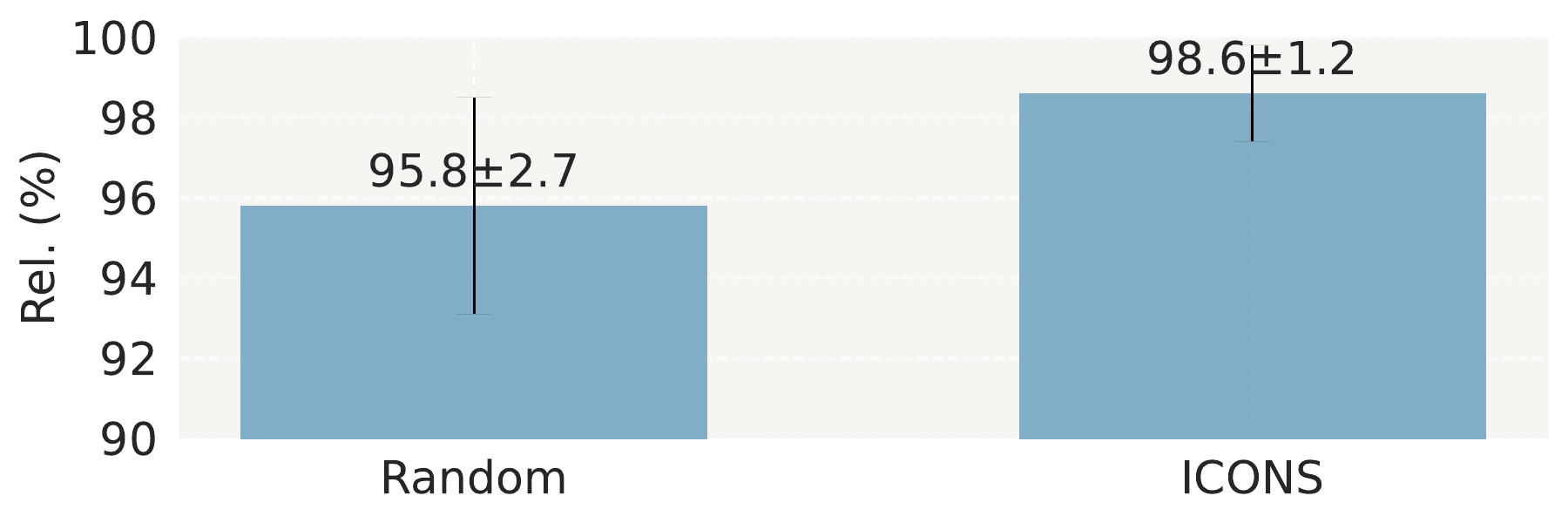}
    %\vspace{-13pt}
    \caption{
    \textbf{Rel. (\%) Across Runs.}
    We show the performance across three different runs for random selection and our \methodname. 
    }
    %\vspace{-10pt}
    \label{fig:consistency}
\end{figure}

\subsection{Consistency analysis}
\label{subsec:consistency}
To evaluate the consistency of \methodname, we conduct three independent runs of the experiments. 
As shown in Fig.~\ref{fig:consistency}, our method demonstrates high consistency across different runs, achieving 98.6$\pm$1.2\% \relp, which shows a notable improvement over random baseline, which achieves 95.8$\pm$2.7\%. 
The lower standard deviation (1.2\% vs 2.7\%) further indicates that our approach produces more stable and reliable outcomes compared to the random baseline.

\section{Algorithm details}
\label{sec:algo_details}
We provide detailed pseudocode for our two-stage ICONS framework. Stage \ref{alg:icons-specialist} (specialist) computes task-specific influence scores through gradient-based analysis with efficient random projections.  Stage~\ref{alg:icons-generalist} (generalist) implements our voting-based consensus mechanism to select samples that are influential across multiple tasks. 

\begin{algorithm}[h]
\caption{ICONS Stage 1: Specialist (Task-specific Influence Computation)}
\label{alg:icons-specialist}
\begin{algorithmic}[1]
\REQUIRE Training dataset $D = \{(x_i, I_i, y_i)\}_{i=1}^N$, target tasks $T = \{T_1, ..., T_K\}$
\REQUIRE Warm-up ratio $r$ (default 5\%)
\ENSURE Task-specific influence scores $\{\bar{I}_k\}_{k=1}^K$

\FOR{each task $T_k \in T$}
    \STATE // Step 1: Warm-up Training
    \STATE Sample warm-up set $D_{\text{warmup}} \subset D$ of size $r|D|$
    \STATE $f_{\text{warmup}} \gets \text{LoRA}(f_{\text{base}}, D_{\text{warmup}})$
    
    \STATE // Step 2: Gradient Computation
    \FOR{each training data $z_i \in D$}
        \STATE $g_i \gets \nabla_{\theta_w} L(f_{\text{warmup}}(z_i), y_i)$
        \STATE $\tilde{g}_i \gets \text{Normalize}(Rg_i)$ \COMMENT{Random projection}
    \ENDFOR
    
    \FOR{each validation data $z'_j \in D_{\text{val}}^k$}
        \STATE $g'_j \gets \nabla_{\theta_w} L(f_{\text{warmup}}(z'_j), y'_j)$
        \STATE $\tilde{g}'_j \gets \text{Normalize}(Rg'_j)$
    \ENDFOR
    
    \STATE // Step 3: Influence Matrix Computation
    \FOR{each $z_i \in D$, $z'_j \in D_{\text{val}}^k$}
        \STATE $I_{ij}^k \gets \langle\tilde{g}_i, \tilde{g}'_j\rangle$
    \ENDFOR
    
    \STATE // Compute average influence per training sample
    \STATE $\bar{I}_k(z_i) \gets \frac{1}{|D_{\text{val}}^k|}\sum_{j=1}^{|D_{\text{val}}^k|} I_{ij}^k$
\ENDFOR

\RETURN Task-specific influence scores $\{\bar{I}_k\}_{k=1}^K$
\end{algorithmic}
\end{algorithm}

\begin{algorithm}[h]
\caption{ICONS Stage 2: Generalist (Influence Consensus-based Data Selection)}
\label{alg:icons-generalist}
\begin{algorithmic}[1]
\REQUIRE Task-specific influence scores $\{\bar{I}_k\}_{k=1}^K$
\REQUIRE Selection ratio $p$, number of tasks $K$
\ENSURE Selected subset $S \subset D$ of size $m \ll N$

\STATE // Compute voting thresholds
\FOR{each task $T_k \in T$}
    \STATE $\tau_k \gets (1-p)\text{-th percentile of } \{\bar{I}_k(z_i)\}_{i=1}^N$
\ENDFOR

\STATE // Voting process
\FOR{each training sample $z_i \in D$}
    \STATE $I_{\text{vote}}(z_i) \gets 0$
    \FOR{each task $T_k \in T$}
        \STATE $\text{vote}_k(z_i) \gets \mathbbm{1}[\bar{I}_k(z_i) \geq \tau_k]$
        \STATE $I_{\text{vote}}(z_i) \gets I_{\text{vote}}(z_i) + \text{vote}_k(z_i)$
    \ENDFOR
\ENDFOR

\STATE // Select top samples based on total votes
\STATE $S \gets \text{top-}p \text{ samples by } I_{\text{vote}}$
\RETURN Selected subset $S$
\end{algorithmic}
\end{algorithm}

\section{Future work}
\label{sec:future}
Our work opens several promising research directions for improving vision-language data selection.
While our work focuses specifically on visual instruction tuning data, our influence consensus approach can be naturally extended to other stages of MLLM training, such as alignment stage. 
The majority voting mechanism may under-represent tasks with unique characteristics or those in the long tail, as it prioritizes samples that broadly benefit multiple tasks to build the \textit{main knowledge pool}. This can lead to limited support for specialized tasks or the reinforcement of spurious correlations spanning multiple tasks. 
Future work could explore \textbf{weighted voting mechanisms}, in which tasks are assigned weights based on their relative importance or contribution to overall model performance for more balanced data selection. Additionally, investigating more efficient gradient computation and storage methods would help scale these methods to larger datasets while maintaining strong performance across diverse vision-language tasks.

\section{Visualizations}
\label{sec:visualization}

\begin{figure*}[t]
    \centering
    \includegraphics[width=0.75\linewidth]{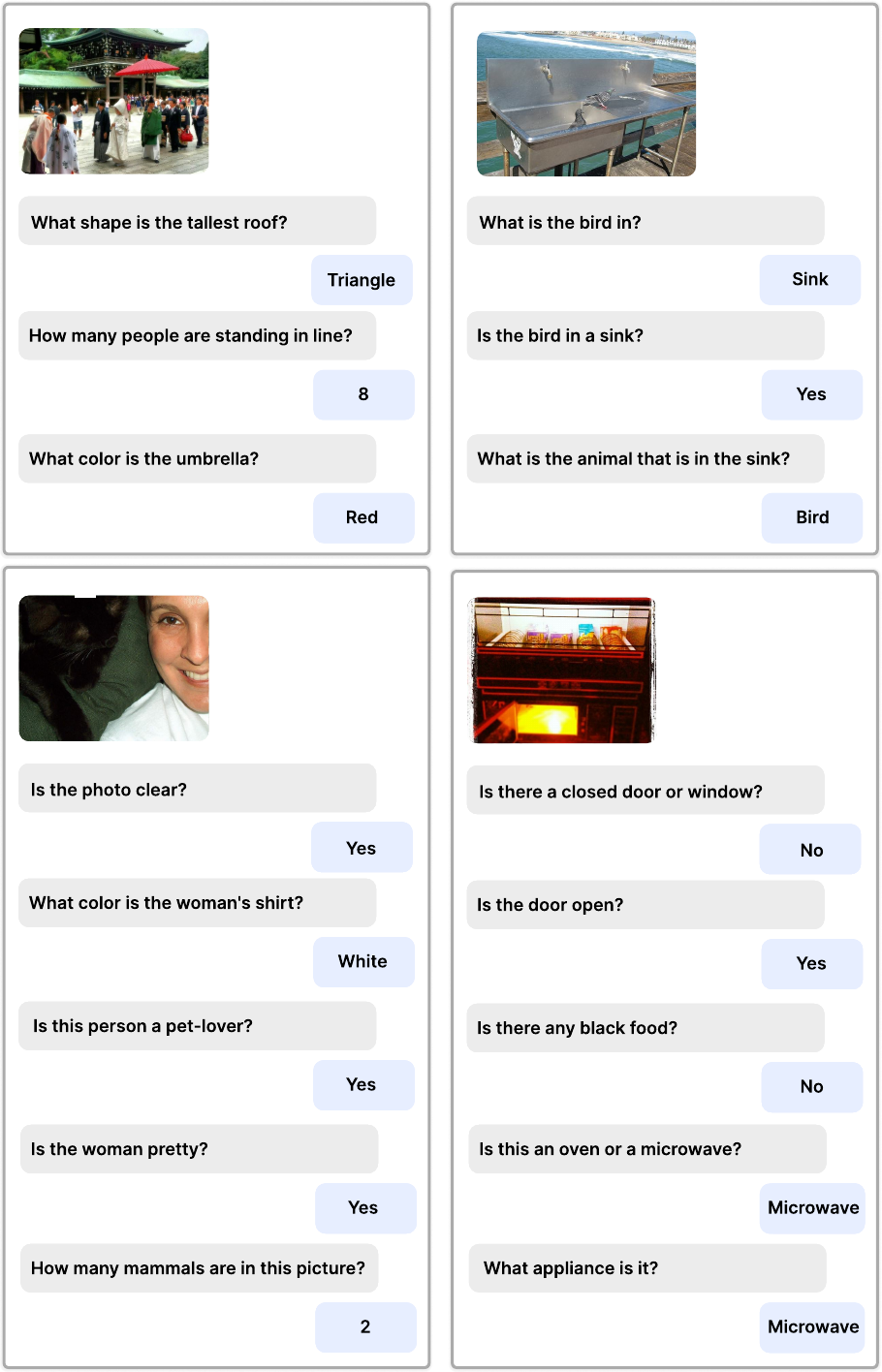}
    \caption{\textbf{Examples of 0 \& 1-vote training samples.} These low-vote samples exhibit characteristics that explain their low selection frequency: (1) subjective questions lacking objective ground truth (e.g., ``Is the woman pretty?''), (2) trivial questions providing minimal visual learning signal (e.g., ``Is the door open?''), (3) redundant questions about the same image content. By filtering out such potentially harmful or redundant examples, \methodname focuses training on more informative and generalizable samples.}
    \label{fig:low_vote_samples}
\end{figure*}

In this section, we provide detailed visualizations. We compare representation-based versus gradient-based data selection approaches (\S\ref{subsec:reps_vs_grads}), showing qualitative differences in selected samples, and visualize the most influential examples identified by both task-specific specialists and the generalist model across all target tasks (\S\ref{subsec:specialist_generalist_vis}). Additionally, we present examples of low-vote training samples (Fig.~\ref{fig:low_vote_samples}), which are referenced in \S\ref{subsec:zero} of the main paper, to illustrate the characteristics of data filtered out by our method.

\subsection{Representation-based vs. Gradient-based data selection}
\label{subsec:reps_vs_grads}

While ICONS leverages gradient-based influence signals, we explore how representation-based data selection (RDS) performs in the same setting (\S\ref{subsec:baseline}). 
We compare the top-ranked training samples selected after the generalist stage by RDS with those selected by \methodname in Fig.~\ref{fig:generalist}.
Interestingly, we observe that the representation-based variants often favor training examples with repeated images or instructions, which may dominate the learned representations without contributing to better generalization. 
Some of the highest-scoring samples under representation-based similarity are duplicated image-question pairs with only the answer choices shuffled. 
We hypothesize that this is a side effect of the way multimodal representations are constructed, where visually dominant or textually redundant samples occupy high-density regions in embedding space. 
However, these samples do not necessarily translate into broader utility across tasks, as seen in the performance gap in Tab.~\ref{tab:main}. 
This discrepancy raises broader questions about what it means for multimodal data to be \textit{diverse}. 
% Should diversity be measured by representation spread, semantic variety, or gradient influence? 
While we leave these questions open for future exploration, our results suggest that gradient-based influence, though computationally more expensive, is better aligned with generalization and multi-task data mixture settings. 

\begin{figure*}[h!]
    \centering
    \includegraphics[width=\textwidth]{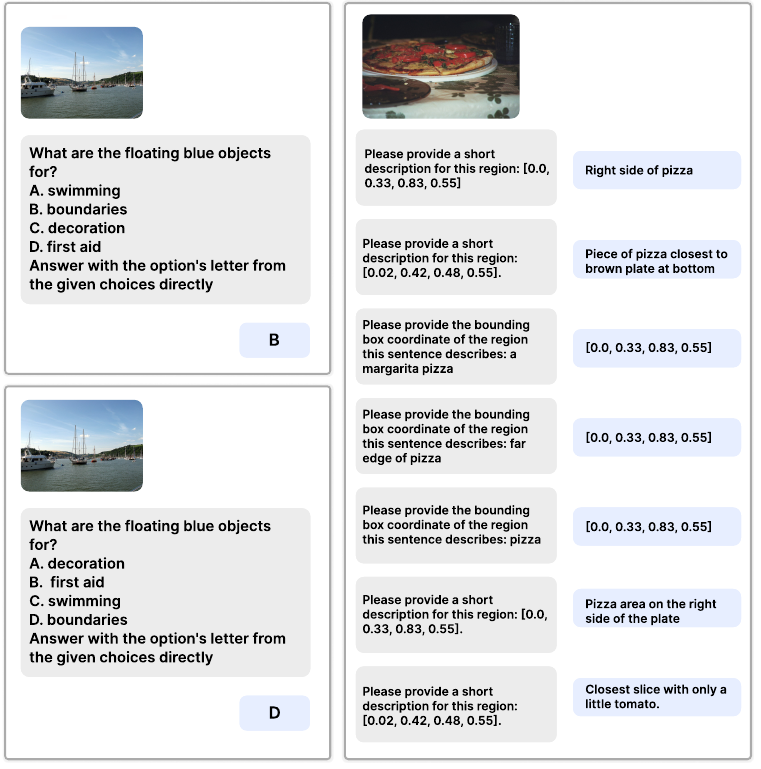}
    %\vspace{-15pt}
    \caption{\textbf{Top three samples selected by RDS.}  These highly-ranked samples are selected due to representation-based similarity but do not necessarily contribute to better generalization, highlighting a key limitation of representation-based selection in multimodal settings.}
     %\vspace{-15pt}
    \label{fig:generalist_reps}
\end{figure*}

\subsection{Specialists \& Generalist}
\label{subsec:specialist_generalist_vis}
\begin{figure*}[h!]
    \centering
    \includegraphics[width=0.7\textwidth]{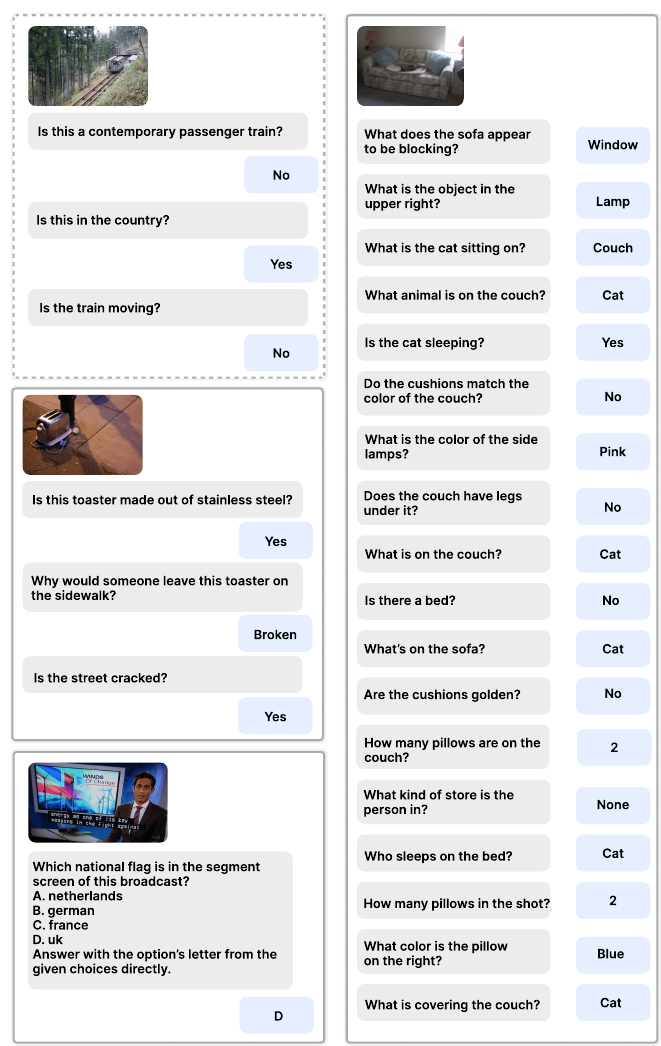}
    \caption{\textbf{VQAv2}. Top-left: A sample from VQAv2~\cite{vqav2}. Remaining panels show top three influential samples selected using the specialist influence ranking step.}
    \label{fig:vqav2}
\end{figure*}

\begin{figure*}[t!]
    \centering
    \includegraphics[width=\textwidth]{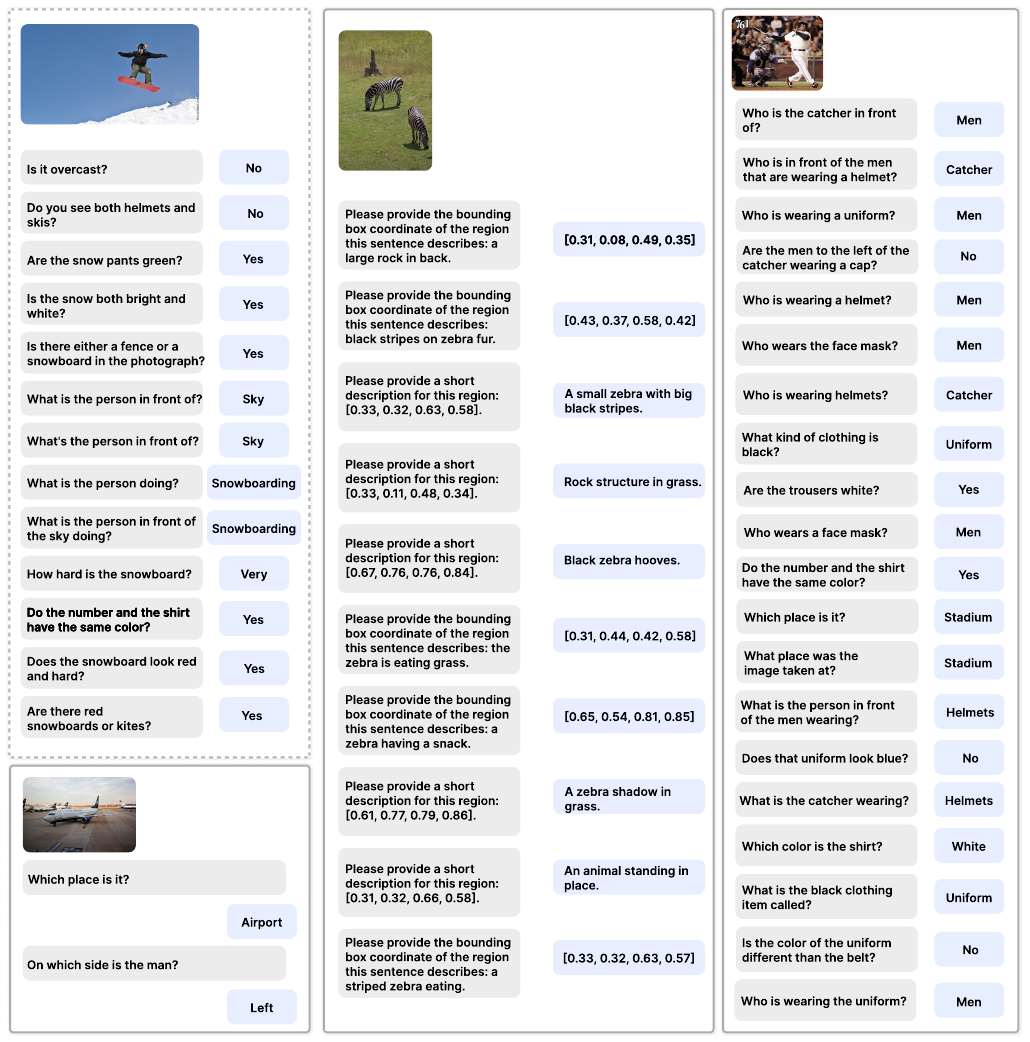}
    \caption{\textbf{GQA}. Top-left: A sample from GQA~\cite{gqa}. Remaining panels show top three influential samples selected using the specialist influence ranking step.}
    \label{fig:gqa}
\end{figure*}

\begin{figure*}[t!]
    \centering
    \includegraphics[width=0.9\textwidth]{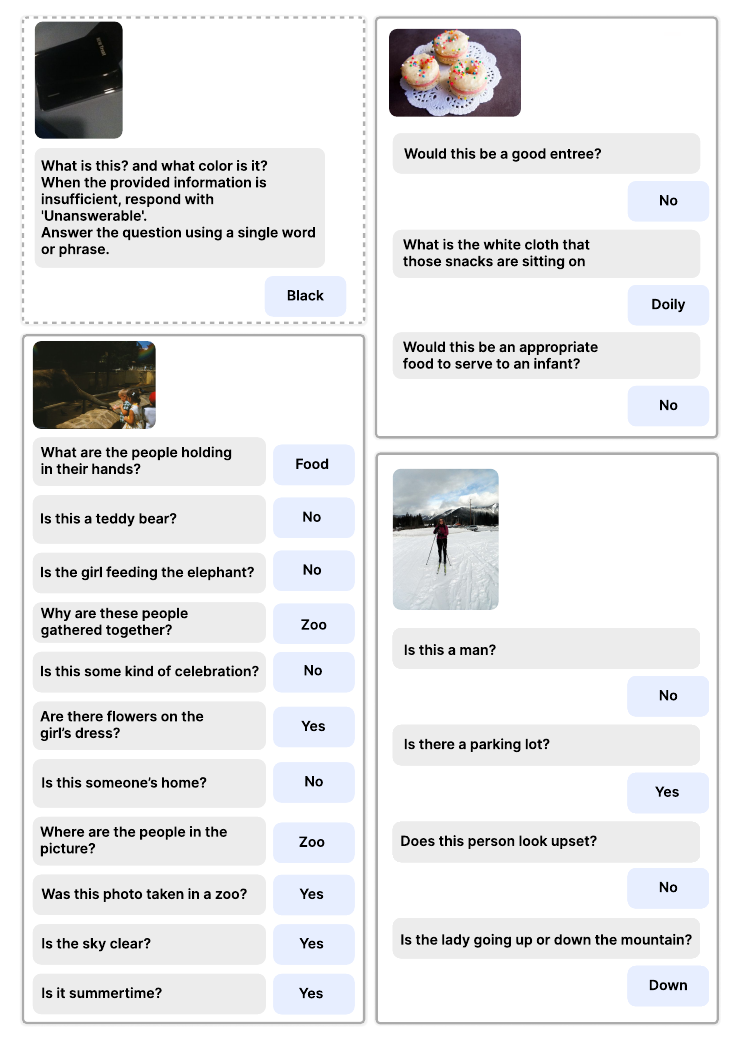}
    \caption{\textbf{VizWiz}. Top-left: A sample from VizWiz~\cite{vizwiz}. Remaining panels show top three influential samples selected using the specialist influence ranking step.}
    \label{fig:vizwiz}
\end{figure*}

\begin{figure*}[t!]
    \centering
    \includegraphics[width=0.8\textwidth]{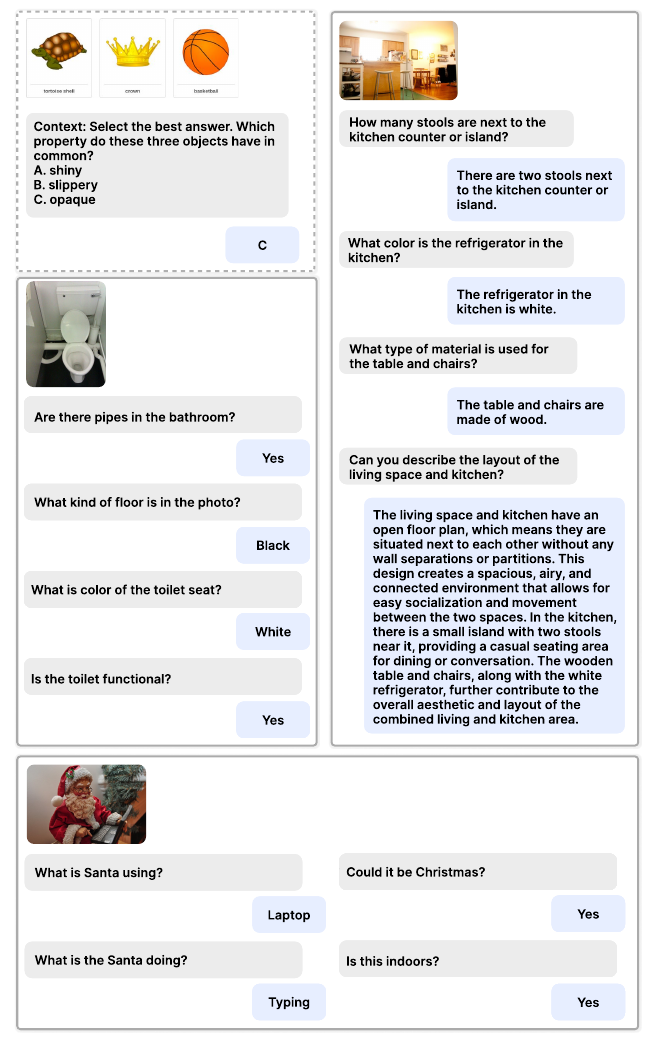}
    \caption{\textbf{SQA}. Top-left: A sample from SQA~\cite{sqa}. Remaining panels show top three influential samples selected using the specialist influence ranking step.}
    \label{fig:sqa}
\end{figure*}

\begin{figure*}[t!]
    \centering
    \includegraphics[width=\textwidth]{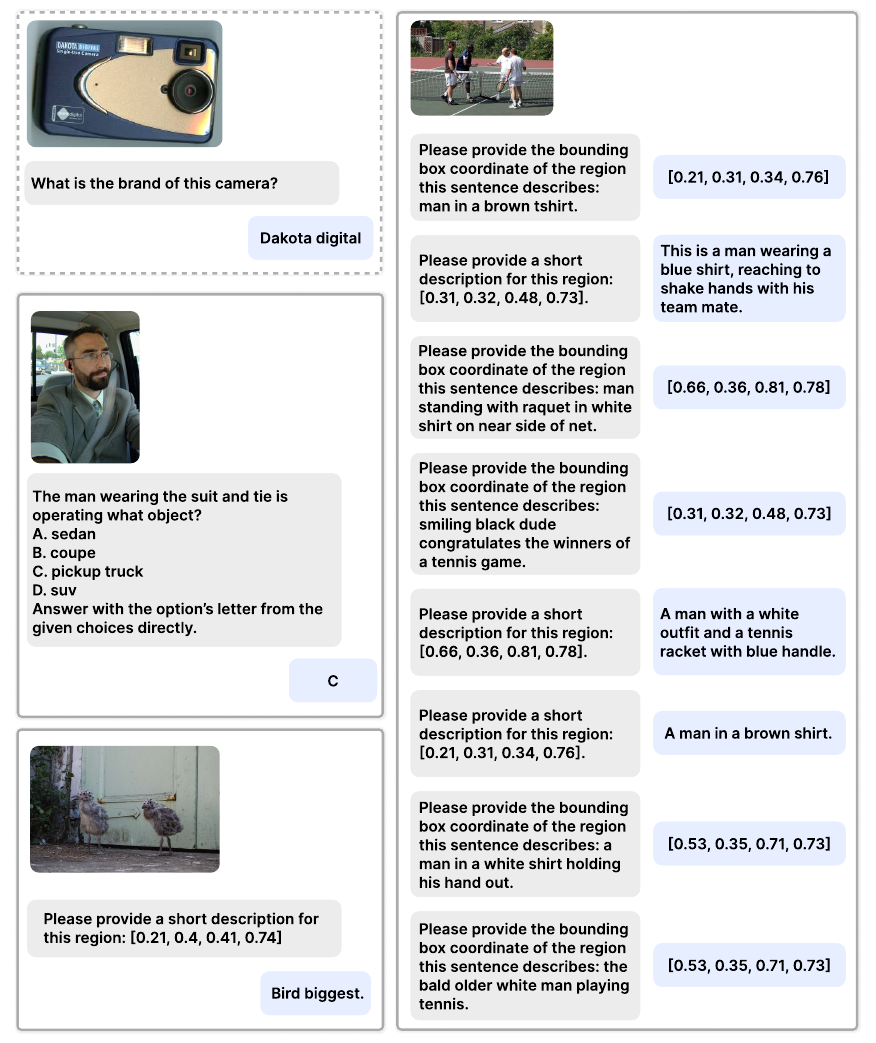}
    \caption{\textbf{TextVQA}. Top-left: A sample from TextVQA~\cite{textvqa}. Remaining panels show top three influential samples selected using the specialist influence ranking step.}
    \label{fig:textvqa}
\end{figure*}

\begin{figure*}[t!]
    \centering
    \includegraphics[width=\textwidth]{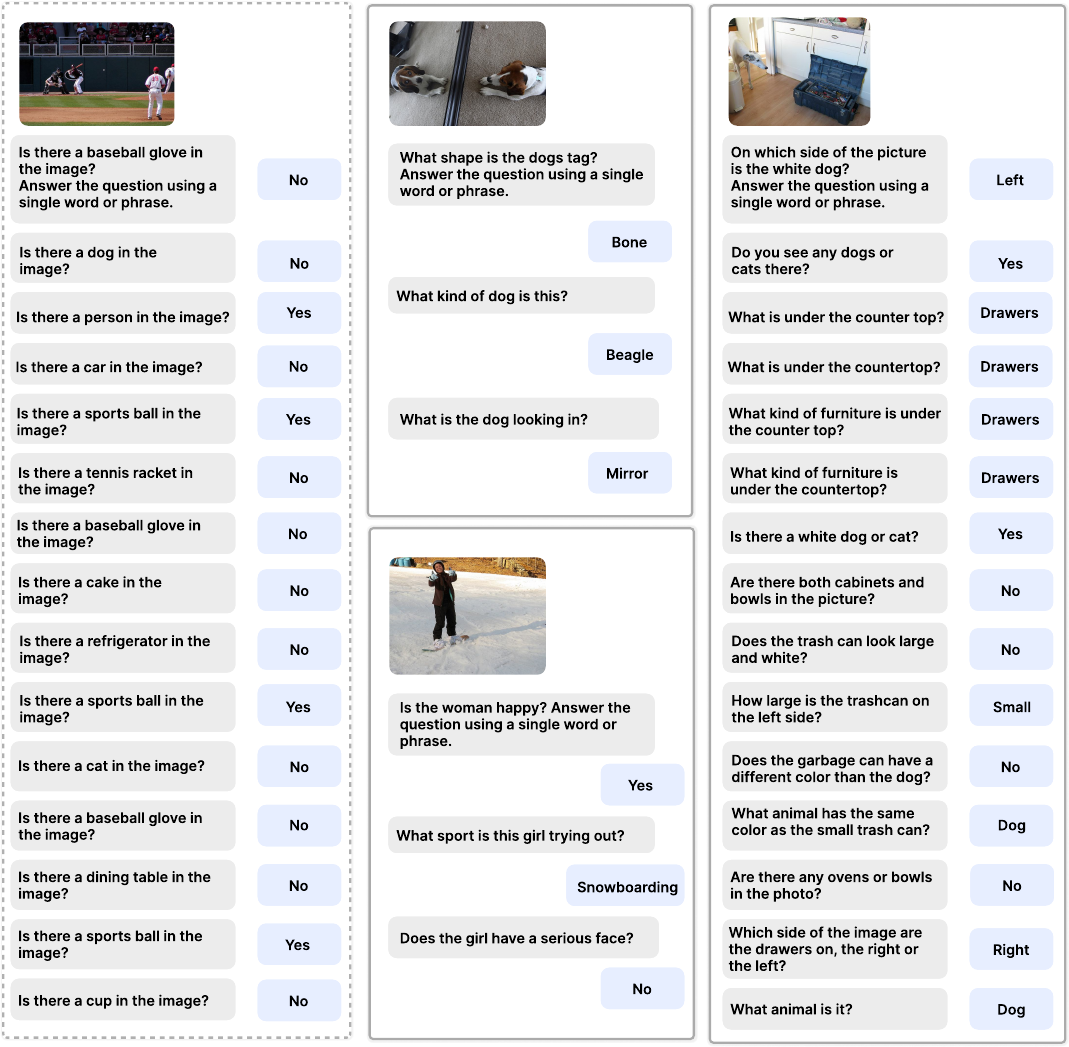}
    \caption{\textbf{Pope}. Top-left: A sample from Pope~\cite{pope}. Remaining panels show top three influential samples selected using the specialist influence ranking step.}
    \label{fig:pope}
\end{figure*}

\begin{figure*}[t!]
    \centering
    \includegraphics[width=0.9\textwidth]{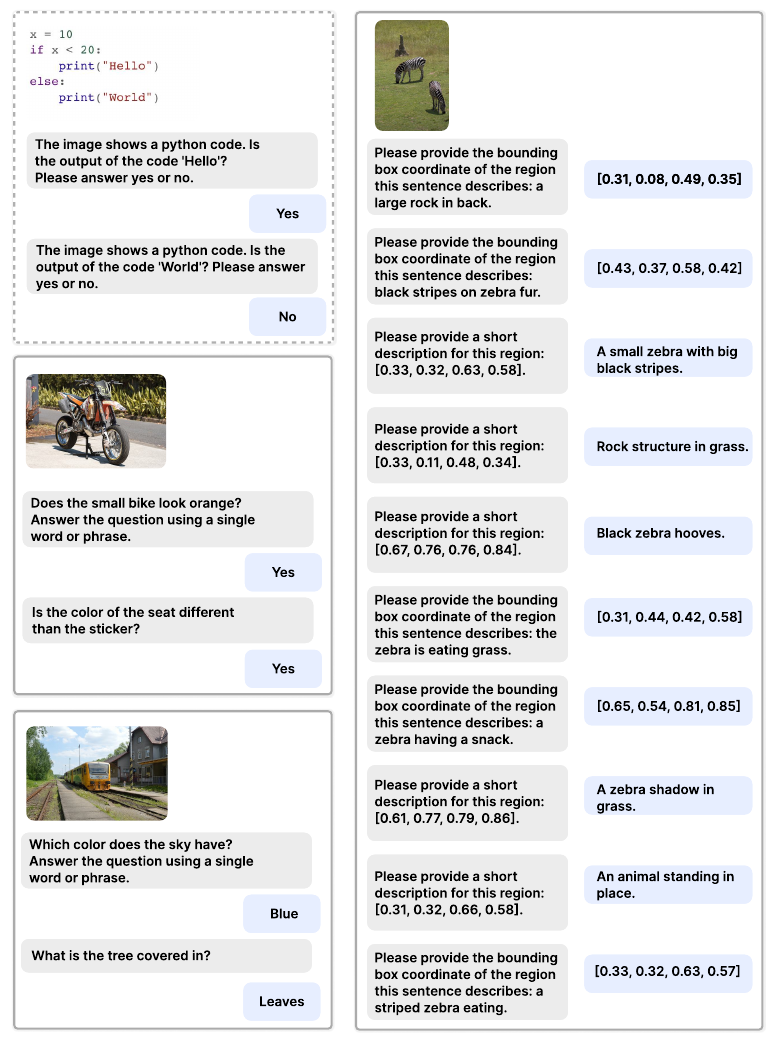}
    \caption{\textbf{MME}. Top-left: A sample from MME~\cite{mme}. Remaining panels show top three influential samples selected using the specialist influence ranking step.}
    \label{fig:mme}
\end{figure*}

\begin{figure*}[t!]
    \centering
    \includegraphics[width=\textwidth]{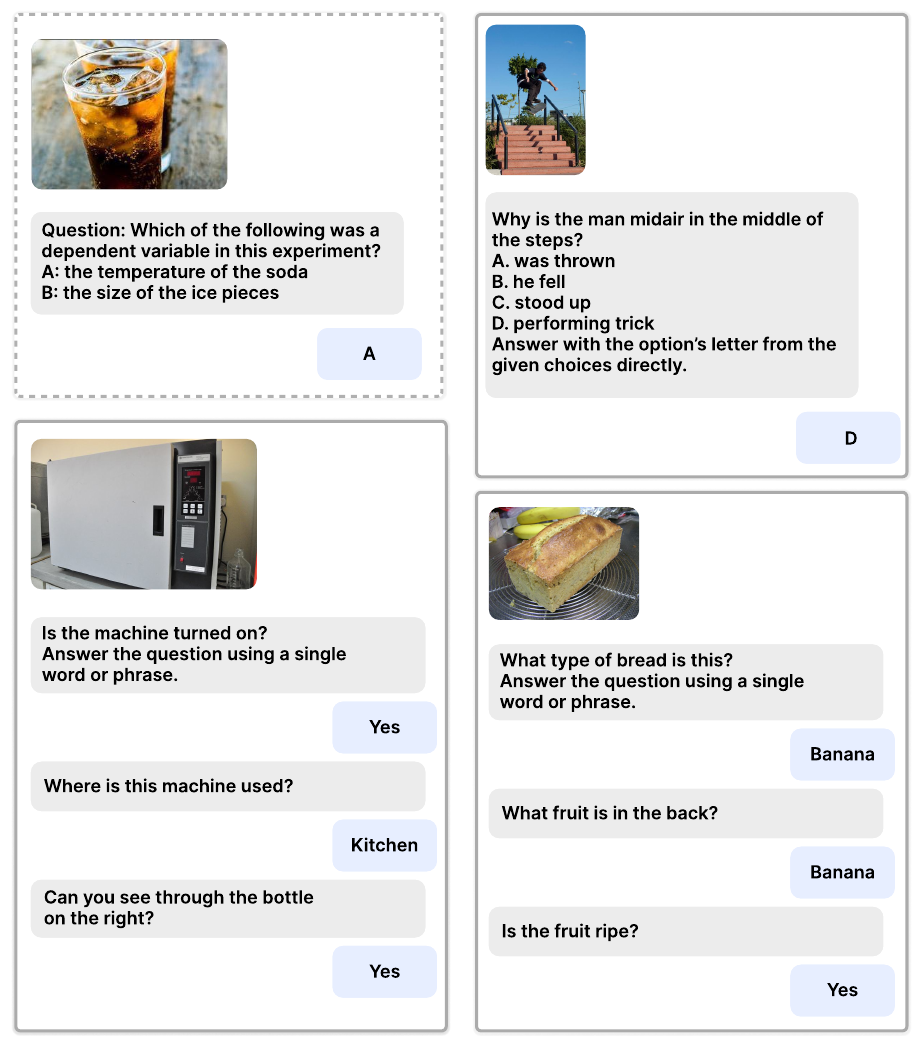}
    \caption{\textbf{MMBench (en)}. Top-left: A sample from MMBench (en)~\cite{mmbench}. Remaining panels show top three influential samples selected using the specialist influence ranking step.}
    \label{fig:mmbench}
\end{figure*}

\begin{figure*}[t!]
    \centering
    \includegraphics[width=0.9\textwidth]{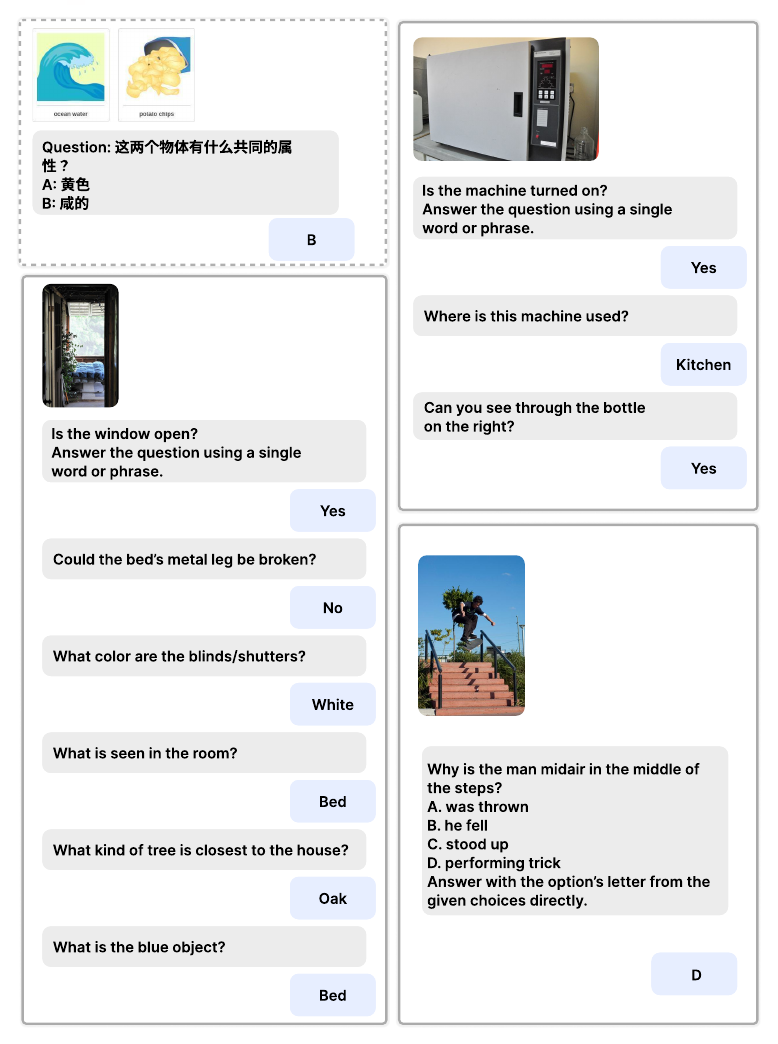}
    \caption{\textbf{MMBench (cn)}. Top-left: A sample from MMBench (cn)~\cite{mmbench}. Remaining panels show top three influential samples selected using the specialist influence ranking step.}
    \label{fig:mmbenchcn}
\end{figure*}

\begin{figure*}[t!]
    \centering
    \includegraphics[width=\textwidth]{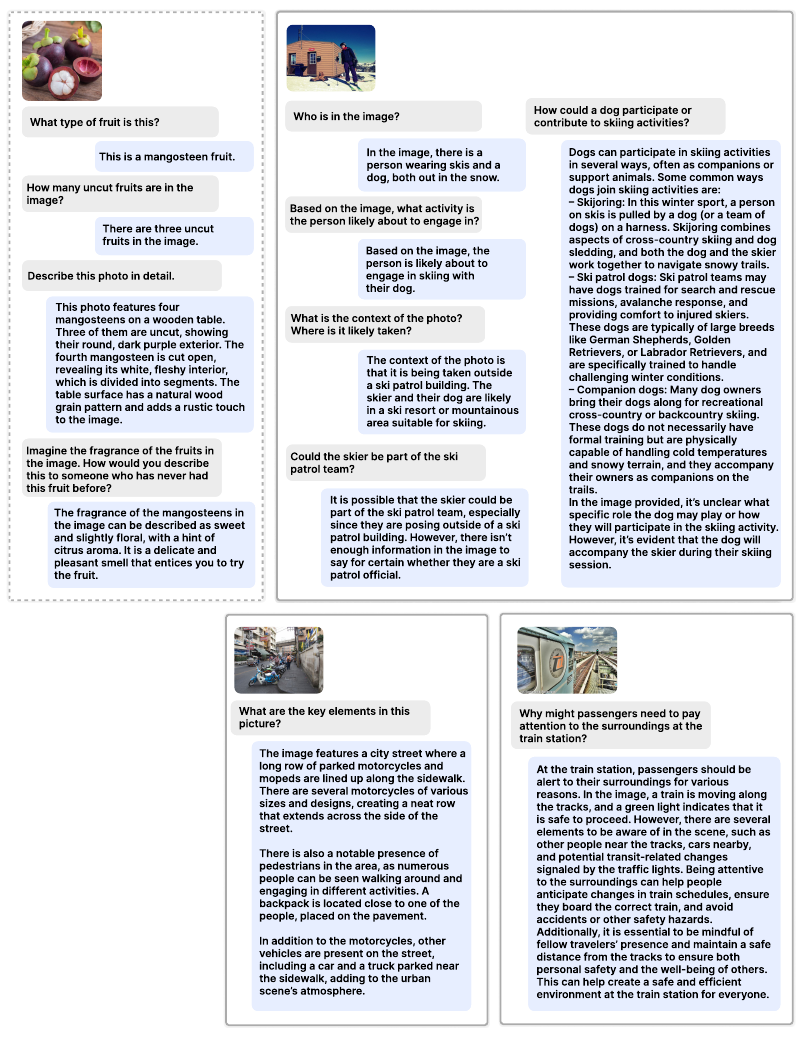}
    \caption{\textbf{LLaVA-W Bench}. Top-left: A sample from LLaVA-W Bench~\cite{llava}. Remaining panels show top three influential samples selected using the specialist influence ranking step.}
    \label{fig:llavaw}
\end{figure*}

\begin{figure*}[t!]
    \centering
    \includegraphics[width=0.7\textwidth]{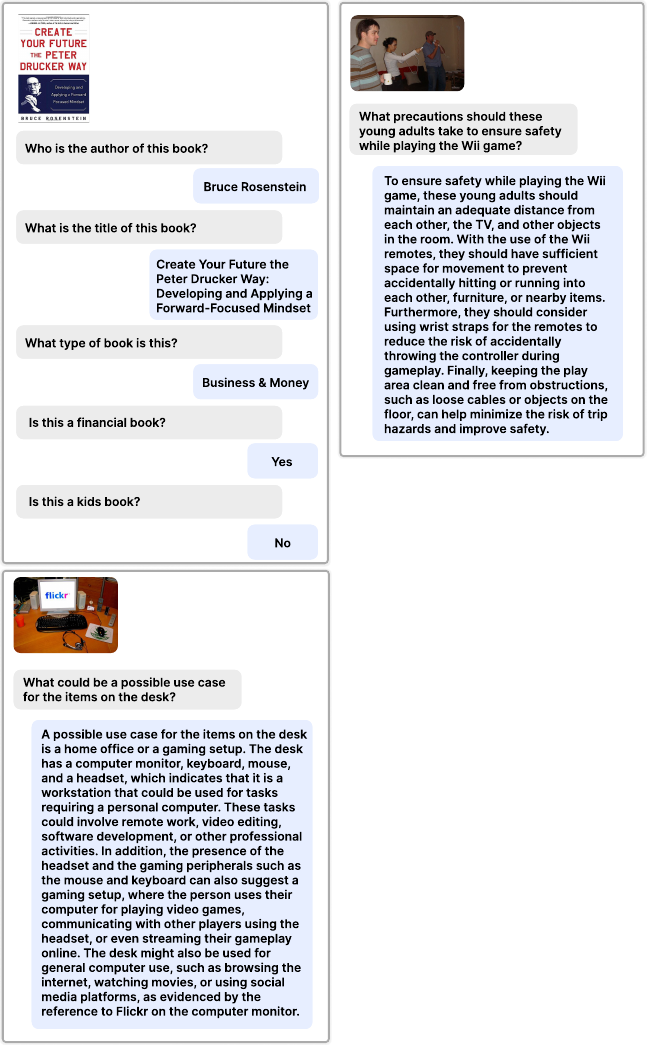}
    \caption{\textbf{Generalist}. We show top three influential samples selected after the generalist stage.}
    \label{fig:generalist}
\end{figure*}

We visualize the most influential top three examples across specialists (\cref{fig:vqav2,fig:gqa,fig:vizwiz,fig:sqa,fig:textvqa,fig:pope,fig:mme,fig:mmbench,fig:mmbenchcn,fig:llavaw}) and the generalist selection (Fig.~\ref{fig:generalist}), along with samples from their corresponding tasks. 
Notably, the selected high-influence examples by specialists show strong task-specific characteristics both structurally and contextually - they mirror the key attributes of their target tasks in terms of question structure, reasoning patterns, and required visual-language understanding capabilities.
Furthermore, the visualization of top influential examples reveals distinct patterns in what makes training samples valuable for different vision-language tasks. VQAv2, GQA, and SQA-I specialists favor multi-turn Q\&A scenarios that test both visual comprehension and contextual understanding, while TextVQA, POPE, and MME specialists emphasize text recognition, object verification, and spatial relationships respectively. MMBench-EN and MMBench-CN show consistent patterns despite language differences, focusing on clear, unambiguous scenes that translate well. The LLaVA-W Bench specialist prioritizes examples requiring detailed explanations and multi-step reasoning, and the answers are generally longer. The generalist model values diverse scenarios that combine multiple skills simultaneously. 
Common characteristics that make these examples particularly valuable include multi-turn interactions, clear visual elements, factual and inferential reasoning, cross-modal interaction, and the ability to test multiple capabilities within a single example. 
This suggests that the most effective training samples are those that combine multiple types of reasoning while maintaining clear, unambiguous ground truth that can be consistently learned across tasks.

\end{document}